\documentclass{article}

\PassOptionsToPackage{numbers, compress}{natbib}

\usepackage[main, final]{neurips_2026}

\usepackage[utf8]{inputenc} 
\usepackage[T1]{fontenc}    
\usepackage{hyperref}       
\hypersetup{hidelinks}
\usepackage{url}            
\usepackage{booktabs}       
\usepackage{siunitx}        
\usepackage{multirow}       
\usepackage{caption}        
\usepackage{amsfonts}       
\usepackage{nicefrac}       
\usepackage{microtype}      
\usepackage[dvipsnames,table]{xcolor}   
\usepackage{adjustbox}

\usepackage{amsmath}
\usepackage{float}
\usepackage[noend]{algpseudocode}
\usepackage{algorithm}
\usepackage{xspace}
\definecolor{keywordcolor}{RGB}{255, 102, 0}   
\definecolor{functioncolor}{RGB}{0, 102, 204}  
\definecolor{operatorcolor}{RGB}{128, 0, 128}  
\newcommand{\kw}[1]{\textbf{#1}}
\newcommand{\fn}[1]{\textcolor{functioncolor}{\text{#1}}}
\newcommand{\op}[1]{\mathrel{\textcolor{operatorcolor}{#1}}}
\algrenewcommand{\algorithmicfor}{\kw{for}}
\algrenewcommand{\algorithmicif}{\kw{if}}
\algrenewcommand{\algorithmicthen}{\kw{:}}
\algrenewcommand{\algorithmicelse}{\kw{else}}
\algrenewcommand{\algorithmicreturn}{\kw{return}}
\algrenewcommand{\algorithmicwhile}{\kw{while}}
\algrenewcommand{\algorithmicdo}{\kw{:}}
\algrenewcommand{\algorithmicrepeat}{\kw{repeat}}
\algrenewcommand{\algorithmicuntil}{\kw{until}}
\usepackage{listings}
\usepackage[skins]{tcolorbox}
\tcbuselibrary{breakable,listings}

\newtcbinputlisting[auto counter, number within=section]{\promptlisting}[4]{%
  enhanced,
  breakable,
  listing only,
  listing file={#3},
  label={#4},
  listing options={
    basicstyle=\scriptsize\ttfamily,
    breaklines=true,
    breakatwhitespace=true,
    columns=flexible,
    keepspaces=true,
    showstringspaces=false,
    extendedchars=true,
    literate={—}{{---}}1 {→}{{$\to$}}1 {×}{{$\times$}}1,
    frame=none
  },
  colback=#1!5!white,
  colframe=#1!75!black,
  coltitle=white,
  fonttitle=\bfseries\small,
  title=Box~\thetcbcounter: #2,
  boxsep=2pt,
  left=4pt, right=4pt, top=2pt, bottom=2pt,
}

\newcommand{\passk}{pass\textasciicircum k\xspace}
\newcommand{\passone}{pass\textasciicircum 1\xspace}
\newcommand{\passthree}{pass\textasciicircum 3\xspace}

\newcommand{\ourbenchmark}{$\tau^{c}$-Bench\xspace}

\newcommand{\taubench}{$\tau^2$-Bench\xspace}
\newcommand{\taubenchverified}{$\tau$-Bench-Verified\space}
\newcommand{\tbv}{$\tau$BV\xspace}

\title{A Matter of TASTE: Improving Coverage and Difficulty of Agent Benchmarks}

\author{
  Tomer Keren \\
  Faculty of Data and Decision Science\\
  Technion\\
  \texttt{tkeren@campus.technion.ac.il}
  \And
  Nitay Calderon \\
  Faculty of Data and Decision Science\\
  Technion\\
  \texttt{nitay@campus.technion.ac.il}
  \And
  Asaf Yehudai \\
  IBM Research \\
  \texttt{Asaf.Yehudai@ibm.com}
  \And
  Yotam Perlitz \\
  IBM Research \\
  \texttt{yotam.perlitz@ibm.com}
  \And
  Michal Shmueli-Scheuer \\
  IBM Research \\
  \texttt{shmueli@il.ibm.com}
  \And
  Roi Reichart \\
  Faculty of Data and Decision Science\\
  Technion\\
  \texttt{roiri@technion.ac.il}
}

\begin{document}
\maketitle
\begin{abstract}
As agent capabilities advance, existing benchmarks, such as $\tau^2$-Bench, are becoming increasingly saturated. Yet constructing new benchmark tasks remains complex, costly, and labor-intensive. Moreover, the standard approach, in which scenarios are first written in natural language and then mapped to tool sequences, captures only a narrow subset of the tool-use patterns agents exercise. In this paper, we address these problems by reversing the task construction process. We propose TASTE: Task Synthesis from Tool Sequence Evolution, an automatic method that generates challenging tasks with broader tool-use coverage. TASTE utilizes an Adaptive Contrastive $n$-gram model trained on LLM-judged validity signals. This enables sampling valid tool sequences that cover a vast range of tool combinations. TASTE then selects representative sequences from the pool via clustering, instantiates them into complete benchmark tasks, and refines them through iterative difficulty evolution. Using TASTE, we construct $\tau^c$-Bench, a challenging extension of the three domains of $\tau^2$-Bench. We evaluate $11$ agent/user LLM pairs and find that models nearly saturating $\tau^2$-Bench suffer severe performance drops on our tasks (e.g., Gemini-3-Flash falls from $0.82\!-\!0.94$ to $0.28\!-\!0.61$). Beyond increasing difficulty, our generated tasks more than double the number of unique tool combinations agents must execute. Our results suggest high scores on existing benchmarks often reflect saturation rather than robust task-solving ability. By automating the generation of difficult, high-coverage benchmarks, TASTE enables continuous, scalable evaluation of future agents.\footnote{The code for TASTE and the $\tau^c$-Bench data are available at \href{https://github.com/tomerkeren42/TASTE-task-synthesis-from-tool-sequence-evolution}{GitHub repo}.}
\end{abstract}

\section{Introduction}

Tool-using agents interact with and modify external environments through tool invocations \citep{toolformer23shick}, such as executing code \citep{sweagent24yang, skyrl25cao}, legal complaint drafting, managing customer accounts, or placing orders \citep{lmars25wang, crmweaver25lai, fama25yan}. Unlike standard input-output systems, tool-using agents operate through multi-step actions that evolve the world state, making evaluation dependent not on any single response but on the cumulative effect of the entire interaction \citep{yehudai2025surveyevaluationllmbasedagents}. A natural evaluation paradigm, adopted by benchmarks such as $\tau$-Bench \citep{taubench_24_yao, tau2barres}, is final-state evaluation: a task is solved if the world state at termination matches a target state induced by a gold tool-call sequence.
Constructing benchmarks under this paradigm is particularly demanding. Authoring even a single task involves writing natural-language instructions, configuring the environment, deriving the correct sequence of tool calls, and verifying that the target state is both consistent and reachable \citep{saber2025cuadron, bestpractices2025Zhu}. At the same time, agent capabilities are advancing rapidly, and leading models are approaching the ceiling of existing benchmarks \citep{yehudai2025surveyevaluationllmbasedagents, taubench_24_yao, tau2barres, toolllm24qin, bfcl_25_patil, webarena_24_zhou, swebench_24_Jimenez}.

The combination of costly manual construction and accelerating saturation creates an urgent need for automatic benchmark generation \citep{autocodebench25chou, deepscholarbench25patel}. However, effective automation first requires identifying the properties a good benchmark should satisfy and how to measure them. Our paper begins by proposing three desiderata for benchmarks of tool-using agents. \emph{Validity} requires every task to be both automatically verifiable and correct in that the gold final state is reachable from the task specification. \emph{Difficulty} requires tasks to meaningfully separate agents of different capability levels. \emph{Coverage} requires benchmark tasks to span structurally diverse patterns of tool use rather than oversampling redundant scenarios. Among these three, coverage has received the least attention and lacks a clear operationalization \citep{Mohammadi_survery_2025}. In this paper, we define it through the lens of tool sequences: the ordered list of tool names in the gold trajectory leading to the target state. Framing coverage in terms of tool sequences also opens a constructive pathway: if the space of valid tool sequences can be explored and sampled directly, it becomes possible to design an automatic method for task generation that explicitly targets all three desiderata.

\begin{figure*}[!t]
    \centering
\includegraphics[width=0.98\textwidth]{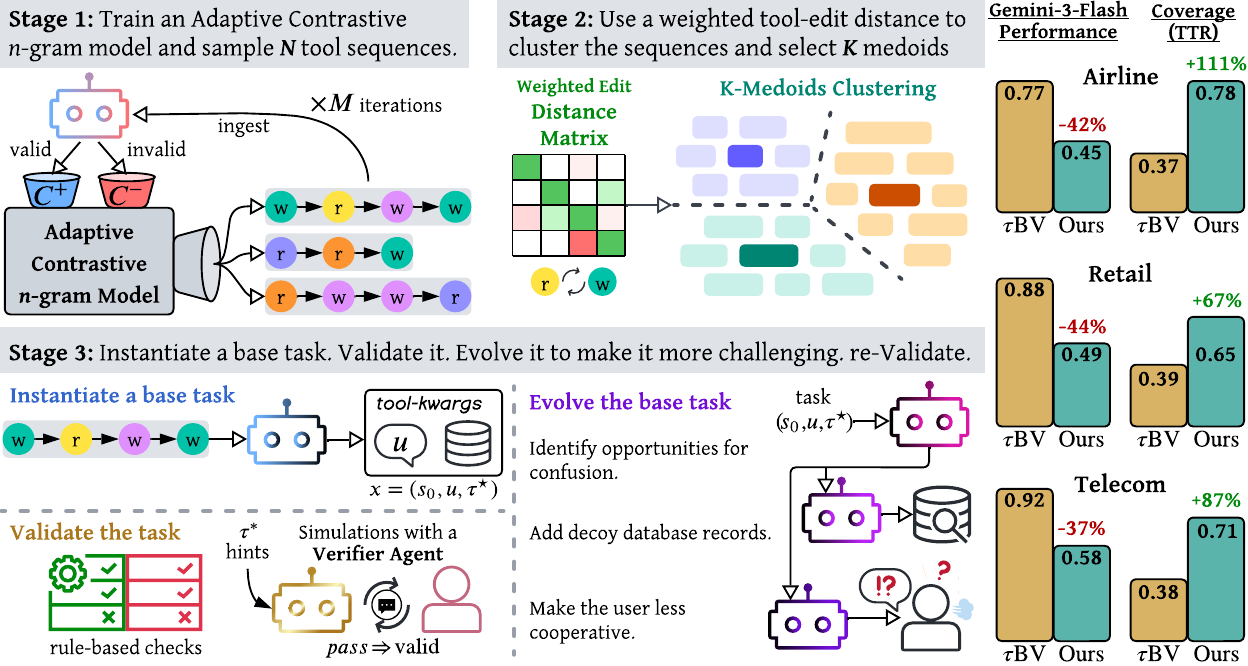}
    \caption{Illustration of TASTE, our three-stage method for synthetic task generation. We demonstrate the method across the three domains of $\tau^2$-Bench (Verified, $\tau\text{BV}$). Evaluating a Gemini-3-Flash agent (averaged over two different user LLMs) shows that, unlike $\tau\text{BV}$, \ourbenchmark is non-saturated. Besides being more challenging, \ourbenchmark exhibits broader coverage, requiring agents to exercise a more diverse set of tool combinations, as quantified by the type-token ratio (TTR; see \S\ref{sec:results}). }
    \label{fig:introfig}
    \vspace{-0.8em}
\end{figure*}

We propose to reverse the task construction process entirely. The prevailing approach involves an author who first writes a natural-language scenario and then derives the corresponding tool sequence. Because the tool-sequence space is never explicitly explored, the resulting benchmarks exercise only arbitrary tool combinations that vary with author choice \citep{Mohammadi_survery_2025}. Accordingly, rather than starting from scenarios and recovering tool sequences, we sample a diverse set of tool sequences and synthesize the full tasks around them. We instantiate this idea and propose \emph{TASTE (\textbf{Ta}sk \textbf{S}ynthesis from \textbf{T}ool Sequence \textbf{E}volution)}, an automatic method for task generation comprising three stages, each targeting one or more of these desiderata. TASTE is illustrated in Figure~\ref{fig:introfig}.

In the first stage, we train an Adaptive Contrastive $n$-gram model over the space of tool sequences. The model maintains separate count tables for $n$-grams (units are tools) observed in plausibly valid and implausibly valid sequences, as judged by an LLM, and computes sampling probabilities from their contrastive ratio. Counts are updated online through iterative sample-and-validate loops, progressively refining the distribution. This targets \emph{validity} by concentrating probability mass on valid tool sequences, and \emph{coverage} by exploring regions of the tool-combination space.
The second stage selects a representative subset of $K$ (the benchmark size) sequences from the sampled pool using $K$-medoids clustering \citep{kmedoids}. We employ a weighted Levenshtein distance that reflects semantic and functional similarity between tools.
In the third stage, each selected tool sequence is instantiated into a complete benchmark task (user instruction, database records, tool arguments).
Following rule-based checks, we employ a hint-assisted verifier agent to ensure the task is actually solvable: we provide a corrupted version of the gold tool-call sequence (tools are shuffled, arguments are masked), requiring the agent to cooperate with the simulated user to reach the target state.
A final evolution step targets \emph{interaction difficulty}: making the user ambiguous and less cooperative, and introducing confusing decoy records to the database (e.g., a fully booked flight on the desired route).

Using TASTE, we construct \ourbenchmark, a challenging extension of the three domains of \taubench \citep{tau2barres}: Airline, Retail, and Telecom ($K\!=50\!/114\!/114$ respectively). We then evaluate 11 agent/user LLM pairs across all three domains. We find performance on our \ourbenchmark is consistently and substantially lower than on the original \taubench, with agents scoring up to $80\%$ worse. The degradation is particularly striking for models that appear to have nearly saturated the \taubench: Gemini-3-Flash achieves $0.82\!-\!0.94$ on the original yet drops to $0.28\!-\!0.61$ on ours. Beyond increased difficulty, our benchmark more than doubles the number of unique tool combinations agents must execute: the weighted edit distance between sequences rises by up to $124\%$, the type-token ratio (TTR, the proportion of unique $n$-grams) by up to $111\%$, and tool-frequency entropy by $35\%$. These results suggest that high scores on existing benchmarks often reflect saturation rather than robust task-solving ability, underscoring the need for automated methods like \ourbenchmark that can continuously generate harder tasks with broader coverage.

We additionally adapt TASTE to AppWorld~\mbox{
\citep{appworld24trivedi}}\hskip0pt
, where an autonomous agent executes Python code over application APIs without a simulated user. This extension retains the sequence-first construction framework while adapting task realization, difficulty evolution, and evaluation to a different interaction setting (Section~\ref{sub:appworld}).

To summarize, our contributions are as follows:
(1) We articulate three desiderata for benchmarks of tool-using agents: \emph{validity}, \emph{difficulty}, and \emph{coverage}. We further operationalize coverage through tool sequences.
(2) We propose TASTE, a method for automatic task generation that enables exploration of the tool-sequence space.
(3) We construct \ourbenchmark, a challenging extension of \taubench, evaluate it on 11 agent/user LLM pairs, and demonstrate that \ourbenchmark is harder and provides broader coverage of tool-use patterns.

\section{Evaluating Conversational Tool-Using Agents}
\label{sec:evaluation}

We focus on conversational tool-using agents, a particularly complex and interesting special case of general tool use, in which static, single-turn prompts are replaced by continuous interactions between two agents. This setup naturally subsumes single-turn scenarios.
Our setup involves a conversation between two agents: the evaluated agent and the simulated user. The interaction begins with the evaluated agent receiving a task. At each turn, an agent takes an \emph{action}, either sending a message to the other agent or performing a \emph{tool call}. If the action is sending a message, the turn passes to the other agent. If the action is a tool call, the environment may modify the world state or return an observation, visible only to the calling agent, after which the agent continues its turn. This setup makes evaluation fundamentally more challenging than in standard input-output systems. Evaluation must be carried out over trajectories of actions that are shaped jointly by the agents and the environment, and also evolve the world state \citep{yehudai2025surveyevaluationllmbasedagents}. As a result, correctness must be determined by the resulting world state. A natural solution adopted by the $\tau\text{-Bench}$ benchmark \citep{taubench_24_yao} is \emph{final-state evaluation}:\footnote{Final-state evaluation reduces assessment to a single binary check on the end state, overlooking trajectory-level properties such as efficiency, intermediate rewards, and interaction quality. $\tau\text{-Bench}$ supports additional rewards, such as \emph{action checks} or \emph{communication checks}, but in practice these are excluded from the main reward in most tasks. We adopt the final-state reward alone, both for simplicity and for consistency with how results are standardly reported in $\tau\text{-Bench}$ evaluations.} a task is solved if the interaction terminates in a world state equivalent to a predefined \emph{target state}, induced by a gold tool-call sequence. Accordingly, many different trajectories may be correct, as long as they lead to this target state.

Formally, let $E$ denote the environment, which comprises the domain policy (e.g., business logic, guidelines) state space, the set of available tools $\mathcal{T}$, and an execution engine that operationalizes these tools.
A \emph{tool call} is an action $a$ that invokes a tool $t\in\mathcal{T}$ with arguments: $a=t(\texttt{**kwargs})$. A task instance is defined as $x = (s_0, u, \sigma^\star)$, where $s_0$ is the initial state (e.g., a database), $u$ is the user instruction, and $\sigma^\star=(a_1^\star,\ldots,a_{n}^\star)$ is a gold (but not necessarily unique) tool-call sequence. The target final state is then derived as $s^\star = \texttt{apply}(\sigma^\star, s_0)$.
Let $\sigma = (a_1, \dots, a_k)$ denote the tool-call sequence produced during the interaction, i.e., the ordered list of all tool calls made by both agents.
The \emph{reward} is defined based on the state induced by $\sigma$ and $s^\star$: $r_x(\sigma) = \mathbf{1}[\texttt{apply}(\sigma, s_0) \equiv s^\star]$. Finally, we distinguish between a \emph{tool-call sequence} $\sigma$ and its corresponding \emph{tool sequence} $\bar\sigma=(t_1, \ldots, t_k)\in\mathcal{T}^k$, which is its projection onto the tool-identifiers space, that is, the ordered list of which tools were called, stripped of their arguments. In particular, $\bar\sigma^\star$ denotes the gold tool sequence. $\bar\sigma^\star$ is a central concept in our study, as it captures the structural requirements of a task and directly connects to the coverage and difficulty properties of a benchmark.

\subsection{Desiderata for Agent Benchmarks}
\label{sec:benchmark_quality}

We next propose three properties that characterize the quality of a benchmark (i.e., a set of tasks) for tool-using agents.

\paragraph{Validity.} A task is \emph{valid} if it is both \emph{verifiable} and \emph{correct}. Verifiability means that task success can be automatically checked (e.g., via final-state evaluation) rather than relying on subjective judgments. Correctness means that the gold final state $s^\star$ is consistent and faithful with the task specification $(s_0, u)$, i.e., it can be achieved given $s_0$ and a user instructed with $u$. Without a high proportion of valid tasks, agents may be penalized for benchmark defects rather than true capability limitations \citep{bestpractices2025Zhu}. Indeed, \citep{saber2025cuadron} identified dozens of invalid tasks in $\tau\text{-Bench}$, which were corrected in $\tau\text{-Bench Verified}$.

\paragraph{Coverage.}
A useful benchmark should cover the relevant task space, rather than sampling similar tasks for a smaller subspace. In our setting, each task $x$ is associated with a gold tool sequence $\bar\sigma^\star$ (i.e., the names of the tools leading to $s^\star$). We define coverage in terms of these gold tool sequences, thereby abstracting away from superficial linguistic variation in user instructions and initial states, and focusing instead on the procedural structure of the task. A benchmark has good coverage if its tasks represent structurally different tool sequences, rather than many near-duplicates. In addition, benchmark tasks should reflect common patterns of tool use, rather than arbitrary combinations or isolated outliers. Coverage can be quantified, for example, using the tool-edit distance between sequences or the number of unique $n$-grams of tools. This mirrors standard diversity/novelty quantification in NLP \citep{raven2023mccoy}, with the key distinction that our units are tool names rather than characters or words.

 \paragraph{Difficulty.}
A benchmark should contain tasks that meaningfully differentiate between capability levels. If all tasks are too easy, the benchmark quickly saturates; if all tasks are too hard, it cannot provide a useful ranking of agents. In our setting, task difficulty arises from two sources. \emph{structural difficulty}, determined by properties of the gold tool sequence, such as its length and the types of tools involved (e.g., write operations versus simple reads), and \emph{interaction difficulty}, determined by the instruction ambiguity, initial-state complexity (e.g., distractors, dependencies, stale information), and user behavior (e.g., cooperative vs. adversarial).

\section{TASTE: Task Synthesis from Tool Sequence Evolution}
\label{sec:method}

We next propose TASTE, a synthetic data generation pipeline for constructing new tasks within a given environment. TASTE is designed to satisfy the three properties defined in Section~\ref{sec:benchmark_quality}, enabling fully automated task generation without costly manual curation or correction. It operates around tool sequences, and proceeds in three stages, as illustrated in Figure~\ref{fig:introfig}:

\begin{enumerate}
\item \textbf{Tool Sequence Sampling} (targets \emph{validity} and \emph{coverage}):
We propose a novel sampling method based on an Adaptive Contrastive $n$-gram language model, which is trained to model the distribution of plausible tool sequences. The model samples thousands of diverse candidate sequences under desired constraints, such as the distribution of sequence lengths.

\item \textbf{Clustering and Selection} (targets \emph{coverage}):
We cluster candidate tool sequences using $K$-medoids with a semantically weighted Levenshtein distance. This yields internally coherent clusters of sequences with similar procedural structure, while medoids across clusters capture structurally distinct patterns.

\item \textbf{Task Generation and Evolution} (targets \emph{validity} and \emph{difficulty}):
Each selected sequence (medoid) is instantiated as a complete benchmark task and validated through rule-based checks and a hint-assisted verifier agent. Once a complete task is obtained, we apply difficulty evolution that progressively increases task complexity.
\end{enumerate}

 \subsection{Tool Sequence Sampling with an Adaptive Contrastive $n$-gram Model}
\label{sub:n_gram}

To address \emph{coverage}, this stage produces a large and diverse pool of candidate tool sequences, from which representative sequences are later selected. We use an $n$-gram language model, in which the probability of the $i$-th tool depends only on the preceding $n{-}1$ tools. We call this model the \emph{Adaptive Contrastive $n$-gram model}, as it extends the standard $n$-gram model in two ways. \textit{Adaptive:} the $n$-gram probabilities are updated online as sequences are sampled and validated. \textit{Contrastive:} the model maintains separate count tables for $n$-grams observed in plausibly valid and implausibly valid sequences, and computes sampling probabilities from their ratio.

\paragraph{Model.} Let $\text{ctx}_i=(t_{i-n+1},\ldots,t_{i-1})$ denote the context of length $n{-}1$ preceding position $i$. We maintain two $n$-gram count tables, $C^+$ and $C^-$, constructed from plausible and implausible sequences, respectively. Each table defines a conditional score over the next tool, given the context. The conditional probability of sampling the $i$-th tool given $\text{ctx}_i$ is defined via a contrastive ratio:
\begin{equation*}
S^{\pm}(t_i \mid \text{ctx}_i)
=
\frac{C^{\pm}(\text{ctx}_i,t_i)+\lambda_0}{\sum_{t' \in \mathcal{T}} C^{\pm}(\text{ctx}_i,t') + |\mathcal{T}|\lambda_0},
\qquad
P(t_i \mid \text{ctx}_i) \propto
\Bigg(\frac{S^+(t_i \mid \text{ctx}_i)}{S^-(t_i \mid \text{ctx}_i)^{\lambda_{\text{neg}}}}\Bigg)^{T(k)}.
\end{equation*}
where $\lambda_0>0$ is a Dirichlet smoothing parameter, $\lambda_{\text{neg}}\ge 0$ controls the influence of negative evidence, and $T(k)$ is a temperature that exponentially decays over training iterations.

\paragraph{Adaptive Training.} The model is trained via an iterative generate-and-validate loop that progressively refines the distribution over tool sequences. We initialize the count tables from a seed set of existing tool sequences. At each iteration, we sample candidate sequences autoregressively from the current model, generating each tool conditioned on the preceding $n{-}1$ tools. Each sampled sequence is validated by a prompted LLM, which assigns a binary plausibility label indicating whether the sequence could arise as a solution to a realistic scenario (Box~\ref{box:validate-action-sequence}).  These binary LLM judgments are the \emph{pseudo-labels} used for training.  This label serves as the learning signal: $n$-grams from plausible sequences update $C^+$; those from implausible ones update $C^-$.

\paragraph{Rationale.}
Our design balances coverage and validity. We use an $n$-gram model because it is simple, scalable, and well-suited to tool sequences, where many constraints are local: the plausibility of the next tool often depends on the tools executed recently. Concretely, our sampler is a temperature-annealed log-odds between two Dirichlet-smoothed $n$-gram models over plausible and implausible sequences, trained online via LLM-judged pseudo-labels. The contrastive formulation uses both positive and negative evidence, favoring $n$-grams observed in plausible sequences while penalizing those associated with implausible ones. Negative evidence is especially useful (Figure~\ref{fig:method_analysis}) because some local patterns strongly indicate invalidity, e.g., modifying a reservation after canceling it. Adaptive online training is needed (Figure~\ref{fig:method_analysis}) because the tool-sequence space is combinatorially large and often only a small supervised set is available.

\subsection{Clustering and Selection of Representative Tool Sequences}
\label{sub:clustering}
At this stage, we can sample candidate tool sequences at scale, enabling us to generate a large and diverse pool of $N\!\gg\!K$ candidates from the trained model, where $K$ is the target benchmark size ($N\!=\!2000, K\in\{50, 114\}$). We control sequence lengths by drawing from a configurable distribution (a skew-normal distribution with $\mu\!=\!7$, maximum length of $15$) and enforce uniqueness through resampling. From this pool, we select $K$ representative tool sequences for task generation.

\paragraph{Clustering.} We use $K$-medoids clustering \citep{pam_kaufman1990, kmedoids}, which partitions the pool into $K$ clusters and selects a representative medoid from each. The algorithm minimizes the total distance between each medoid and its cluster members, so each medoid represents a group of similar candidates. The pool provides diversity; medoid selection ensures that central sequences are chosen rather than outliers.

\paragraph{Distance between Tool Sequences.} Using the standard Levenshtein (edit) distance, which assigns unit cost to all operations, led to unintuitive clusters: sequences that differ in semantically important ways were often grouped together. We therefore employ a weighted variant of the Levenshtein distance, in which substitution costs depend on the semantic relationship between tools. Concretely, replacing tools that perform a similar function (e.g., \texttt{search\_direct\_flight} and \texttt{search\_onestop\_flight}) costs $0.33$; replacing tools of the same type (e.g., both \texttt{read} or both \texttt{write}) cost $0.66$; and substitutions across types, insertions, and deletions cost $1$ (Appendix~\ref{app:method_implementation}). A qualitative analysis of clusters and medoids constructed with standard and weighted   edit  distances is provided in Appendix~\ref{app:wed-qualitative}. As shown, clusters formed under the weighted distance are more coherent, with sequences within each cluster having similar functional structure.

\paragraph{Medoids Validation.} We validate medoids using the plausibility LLM validator from the $n$-gram training stage. We apply the following recovery procedure when a medoid is deemed invalid: Replace the invalid medoid with the next-closest member of the same cluster. If all members of a cluster are invalid, remove them from the pool and re-run $K$-medoids with all currently valid medoids fixed as constraints. Repeat until all $K$ medoids are valid.

\subsection{Generating Tasks from Tool Sequence and Evolving the Difficulty}
\label{sub:task_generation}

Given a set of $K$ representative tool sequences, we proceed in two stages: (i) generating a verifiable and correct \emph{base task} for each sequence, and (ii) evolving the task to increase difficulty while preserving its correctness.

\paragraph{Base Task Generation.}
Each tool sequence $\bar\sigma$ is instantiated into a complete, executable task by generating $s_0$ and $u$. This requires two LLM calls. First, the LLM constructs a coherent real-world scenario with respect to the existing environment $E$, and that motivates $\bar\sigma$, inventing concrete entities (e.g., user names, order IDs) and producing a verbose, step-by-step user instruction (Box~\ref{box:create-user-task}). Second, a follow-up call generates the database entities required to execute the task (Box~\ref{box:generate-db-init}). This process yields a \emph{base task}: an intentionally unambiguous instantiation designed to isolate task validity from difficulty (Example~\ref{fig:full-task-examples})

\paragraph{Validity Checks.}

We validate each generated task using a combination of deterministic and simulation-based checks. First, we apply deterministic rule-based checks to verify basic structural validity and ensure that the gold tool-call sequence executes successfully (e.g., that referenced entities exist in $s_0$ and that tool calls conform to the domain schema). These checks are fast and catch structural errors, but they do not ensure that the task is solvable or that the user instruction $u$ contains sufficient information. We therefore perform a simulation-based solvability check using a \emph{Verifier agent}. The agent attempts to solve $x$ through interaction with the user, while receiving partial hints derived from the gold tool-call sequence: the tool calls are shuffled, and for each call, $\lceil p \cdot |\text{kwargs}| \rceil$ arguments are removed. For example, a masked hint is \texttt{send\_certificate(user\_id=???, amount=???)}. Since these hints are incomplete and unordered, they help guide the agent without allowing it to simply replay the solution. The agent must still infer the intended plan, recover missing arguments, and cooperate with the user; The masking fraction and hint order are configured per validation run (Appendix~\ref{app:verifier-hints}). Success, therefore, provides evidence that the task is solvable. If the generated task fails any check, it is regenerated; after repeated failures, the gold tool-call sequence is replaced by another candidate from the same cluster (Section~\ref{sub:clustering}).
\paragraph{Task Evolution.}
The generated base tasks do not reflect realistic conditions: real-world user requests are ambiguous, changing over time, and sometimes adversarial, and the environment may contain confusing or misleading instances. The evolution stage transforms base tasks into harder variants while preserving the gold tool-call sequence (Figure~  \ref{fig:method_analysis}, right ). For every base task, we proceed with the Evolution process through three LLM calls.
\emph{Strategy analysis} examines each write action to identify opportunities for adversarial confusion, drawing from a manually curated catalog of patterns, e.g., demanding an action that the policy forbids, or the user mistakenly provides incorrect information (see Appendix~\ref{app:adversarial-patterns}). \emph{Environment perturbation} constructs decoy database records, entities similar to gold targets but with subtle disqualifying differences (e.g., a flight on the same route with no available seats), which are merged into $s_0$. \emph{Scenario rewriting} rewrites the user instruction to exploit the adversarial opportunities. Each evolved task is re-validated; if validation fails, progressively simpler evolution variants are attempted before falling back to the base task (Appendix~\ref{app:graduated-fallback}).

\begin{table*}[t]
\centering
\large
\caption{\textbf{Main Results:} Performance of agents on $\tau$-Bench Verified (\texttt{$\tau$BV}) and on our generated tasks (\ourbenchmark), across three domains and two user simulators. For the airline domain, we additionally report \passthree alongside the standard \passone scores. $\Delta\%$ denotes the relative change $\frac{\text{Ours} - \tau\text{BV}}{\tau\text{BV}} \times 100$.}
\label{tab:main-results}
\begin{adjustbox}{width=0.98\textwidth}
\begin{tabular}{l|ccc|ccc|ccc|ccc}
\toprule
\cellcolor{Gray!15} 
& \multicolumn{3}{c}{\cellcolor{Gray!40}\textbf{Airline (\passone)}} 
& \multicolumn{3}{c}{\cellcolor{Gray!40}\textbf{Airline (\passthree)}} 
& \multicolumn{3}{c}{\cellcolor{Gray!40}\textbf{Retail (\passone)}} 
& \multicolumn{3}{c}{\cellcolor{Gray!40}\textbf{Telecom (\passone)}} \\
\cellcolor{Gray!15} \textbf{Agent} 
& \cellcolor{Gray!15} $\tau\text{BV}$ 
& \cellcolor{Gray!15} \textbf{Ours} 
& \cellcolor{Gray!15} $\Delta\%$ 
& \cellcolor{Gray!15} $\tau\text{BV}$ 
& \cellcolor{Gray!15} \textbf{Ours} 
& \cellcolor{Gray!15} $\Delta\%$ 
& \cellcolor{Gray!15} $\tau\text{BV}$ 
& \cellcolor{Gray!15} \textbf{Ours} 
& \cellcolor{Gray!15} $\Delta\%$ 
& \cellcolor{Gray!15} $\tau\text{BV}$ 
& \cellcolor{Gray!15} \textbf{Ours} 
& \cellcolor{Gray!15} $\Delta\%$ \\
\midrule
\multicolumn{13}{l}{\cellcolor{Gray!30}\textit{User:~Gemini-3-flash}} \\
Gemini-3-flash 
& 0.72 & 0.34 & \textcolor{BrickRed}{-52.8} 
& 0.56 & 0.22 & \textcolor{BrickRed}{-60.7} 
& 0.88 & 0.44 & \textcolor{BrickRed}{-50.0} 
& 0.90 & 0.55 & \textcolor{BrickRed}{-38.9} \\
Gemini-2.5-flash 
& 0.58 & 0.21 & \textcolor{BrickRed}{-63.8} 
& 0.40 & 0.10 & \textcolor{BrickRed}{-75.0} 
& 0.79 & 0.36 & \textcolor{BrickRed}{-54.4} 
& 0.35 & 0.27 & \textcolor{BrickRed}{-22.9} \\
GPT-5.2 
& 0.57 & 0.49 & \textcolor{BrickRed}{-14.0} 
& 0.34 & 0.26 & \textcolor{BrickRed}{-23.5} 
& 0.92 & 0.59 & \textcolor{BrickRed}{-35.9} 
& 0.55 & 0.44 & \textcolor{BrickRed}{-20.0} \\
Qwen-32B 
& 0.50 & 0.13 & \textcolor{BrickRed}{-74.0} 
& 0.26 & 0.06 & \textcolor{BrickRed}{-76.9} 
& 0.48 & 0.27 & \textcolor{BrickRed}{-43.8} 
& 0.40 & 0.38 & \textcolor{BrickRed}{-5.0} \\
deepseek-3.1 
& 0.49 & 0.41 & \textcolor{BrickRed}{-16.3} 
& 0.30 & 0.16 & \textcolor{BrickRed}{-46.7} 
& 0.47 & 0.47 & 0.0 
& 0.53 & 0.48 & \textcolor{BrickRed}{-9.4} \\
claude-sonnet-4.6 
& 0.72 & 0.64 & \textcolor{BrickRed}{-11.1} 
& -- & -- & -- 
& 0.67 & 0.54 & \textcolor{BrickRed}{-19.4} 
& 0.81 & 0.72 & \textcolor{BrickRed}{-11.1} \\
\midrule
\multicolumn{13}{l}{\cellcolor{Gray!30}\textit{User:~GPT-5.2}} \\
Gemini-3-flash 
& 0.82 & 0.56 & \textcolor{BrickRed}{-31.7} 
& 0.68 & 0.28 & \textcolor{BrickRed}{-58.8} 
& 0.87 & 0.55 & \textcolor{BrickRed}{-36.8} 
& 0.94 & 0.61 & \textcolor{BrickRed}{-35.1} \\
Gemini-2.5-flash 
& 0.66 & 0.36 & \textcolor{BrickRed}{-45.5} 
& 0.40 & 0.08 & \textcolor{BrickRed}{-80.0} 
& 0.60 & 0.50 & \textcolor{BrickRed}{-16.7} 
& 0.46 & 0.33 & \textcolor{BrickRed}{-28.3} \\
GPT-5.2 
& 0.58 & 0.34 & \textcolor{BrickRed}{-41.4} 
& 0.36 & 0.20 & \textcolor{BrickRed}{-44.4} 
& 0.69 & 0.61 & \textcolor{BrickRed}{-11.6} 
& 0.62 & 0.47 & \textcolor{BrickRed}{-24.2} \\
Qwen-32B 
& 0.36 & 0.10 & \textcolor{BrickRed}{-72.2} 
& 0.22 & 0.06 & \textcolor{BrickRed}{-72.7} 
& 0.56 & 0.39 & \textcolor{BrickRed}{-30.4} 
& 0.28 & 0.31 & \textcolor{ForestGreen}{+10.7} \\
deepseek-3.1 
& 0.48 & 0.36 & \textcolor{BrickRed}{-25.0} 
& 0.24 & 0.18 & \textcolor{BrickRed}{-25.0} 
& 0.67 & 0.54 & \textcolor{BrickRed}{-19.4} 
& 0.80 & 0.61 & \textcolor{BrickRed}{-23.8} \\
\bottomrule
\end{tabular}
\end{adjustbox}
\vspace{-0.8em}
\end{table*}

\section{Experimental Setup}
\label{sec:experiments_setup}

\noindent\textbf{Datasets.} We evaluate on and extend the three \taubench domains~\citep{tau2barres}: Airline, Retail, and Telecom ($N\!=\!50/114/114$), using the \taubenchverified corrected task sets~\citep{saber2025cuadron}. Each domain models a customer-support interaction in which a simulated user issues a request and the agent operates over domain-specific tools under a defined policy. Telecom also provides tools to users. Since Telecom gold sequences encode only \emph{write} actions, we restrict gold tool sequences to only write-type tools when applying TASTE and comparing \ourbenchmark to \taubench.

\smallskip\noindent\textbf{TASTE and \ourbenchmark.}
We provide additional implementation details, including a complete description of the hyperparameters in Appendix~\ref{app:method_implementation} and Table~\ref{tab:hparams}. The prompts are described in Appendix~\ref{app:prompts}.
We apply TASTE to each domain, using the original task counts. We train a trigram model ($n\!=\!3$) for 3,000 iterations and sample unique sequences whose lengths are drawn from a skew-normal distribution with $\mu\!=\!7, \sigma\!=\!5, \alpha\!=\!2$, clipped at a maximum length of $15$. The plausibility validator LLM is based on Gemini-3-Flash. For instantiating tasks, we also use  Gemini-3-Flash, which also serves as the hint-assisted verifier agent. For task evolution, we use Gemini-3-Pro.

\smallskip\noindent\textbf{Evaluated agents.}
We use the \taubench harness with default settings. We evaluate six agent LLMs (Gemini-3-Flash, Gemini-2.5-Flash, GPT-5.2, Qwen-32B, deepseek-3.1, Claude-Sonnet-4.6) paired with two user simulators (Gemini-3-Flash and GPT-5.2). All models use vendor-default thinking/reasoning effort settings; exact details are listed in Appendix~\ref{app:impl-hparams}. Following Section~\ref{sec:evaluation}, we report only \passone scores using final-state reward, and additionally report \passthree for Airline.

\smallskip\noindent\textbf{Costs  (\taubench).} Stages 1 and 2 are cost-efficient, costing approximately $\$10$ per domain. Stage 3 (task generation, evolution, and verification) costs approximately $\$2.50$ per task. The total cost of generating \ourbenchmark is $\$725$. Evaluation costs vary by agent--user pair, with a total cost of $\$520$.

\section{Results}
\label{sec:results}

\paragraph{Main Results.} The performance on our generated tasks is consistently and substantially lower against \taubench (Verified version, \tbv), with relative drops ($\Delta\%$) ranging from $-5\%$ to $-80\%$, depending on the agent and domain (Table~\ref{tab:main-results}). We compared the performance of $11$ agent--user pairs across three domains. The degradation is particularly striking for LLMs that appear to have nearly saturated \tbv: Gemini-$3$-flash, for instance, achieves $0.82$--$0.94$ on \tbv across Airline, Retail, and Telecom domains (except Airline with Gemini-$3$-Flash user), yet drops to $0.28$--$0.61$ on \ourbenchmark. This suggests that the high \tbv scores may reflect benchmark saturation rather than robust task-solving ability. \ourbenchmark exposes gaps that the original benchmark no longer captures, enabling differentiation among state-of-the-art LLMs and supporting continuous, scalable evaluation.

\begin{figure*}[t]
    \centering
    \includegraphics[width=0.55\textwidth]{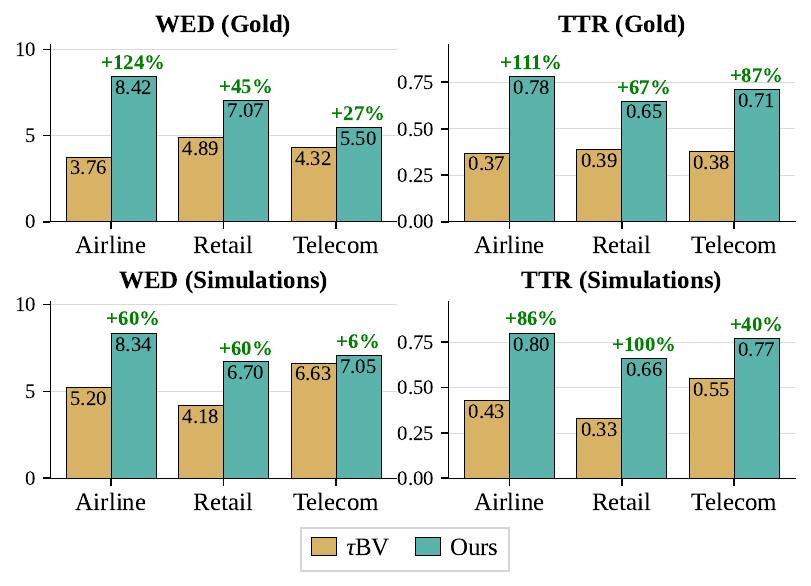}
    \includegraphics[width=0.44\textwidth]{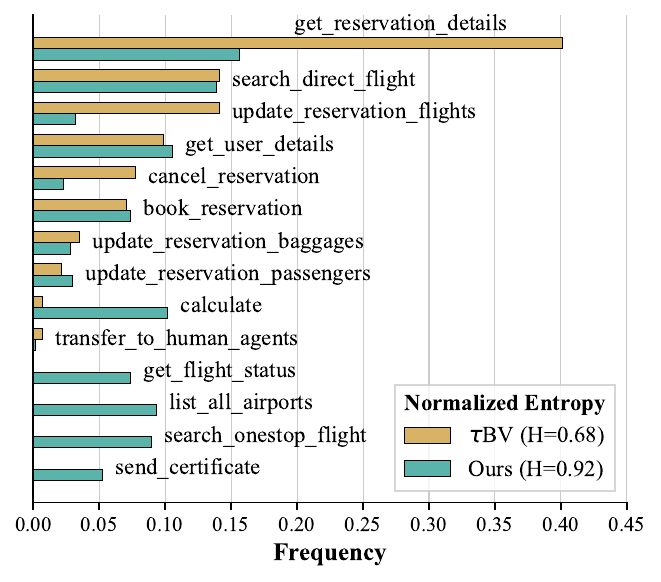}
\caption{\textbf{Coverage Metrics (Left):} Weighted Edit Distance (WED) and Type-Token Ratio (TTR, averaged over $n \in \{2,\dots,6\}$), computed for both gold sequences (top) and sequences extracted from successful simulation trajectories (bottom). \textbf{Airline Tool Frequency (Right):} Distribution of tools in gold sequences. The $\tau\text{BV}$ distribution is notably more skewed. Other domains in Figure~\ref{fig:diversity-combined}.}
    \label{fig:coverage_fig}
\vspace{-1.25em}
\end{figure*}

\paragraph{Harder Tasks with Broader Coverage.}
We compare the coverage and diversity of \ourbenchmark against \tbv, adopting $n$-gram diversity metrics from the NLP literature.
In Figure~\ref{fig:coverage_fig} (left), we present two metrics.
The first is the weighted edit distance (WED) between the tool sequences, capturing the structural dissimilarity of the required execution paths. The second is the Type-Token Ratio (TTR), the fraction of unique $n$-grams ($n\in\{2,...,6\}$) among all $n$-gram occurrences, capturing how much of the available tool-combination space is exercised. We also compute these metrics on tool sequences extracted from successful simulations (see Appendix~\ref{app:sequence-metrics-full}).
Our tasks show consistent and substantial gains across all three domains: WED increases by up to $124\%$, TTR by up to $111\%$,
and the tool-frequency distribution is markedly less skewed with entropy increased by $35\%$ (Figure~\ref{fig:coverage_fig}, right).
These results indicate our tasks require agents to follow more varied and distinct execution paths. A more detailed analysis (Appendix~\ref{app:sequence-metrics-full}) applies additional metrics; across all domains, metrics, and sequence sources, our tasks are consistently more challenging and diverse.

\subsection{Method Analysis}
\label{sub:method_analysis}

In this subsection, we analyze key design choices of TASTE and provide empirical justifications.

\begin{figure*}[t]
    \centering
    \includegraphics[width=0.48\textwidth]{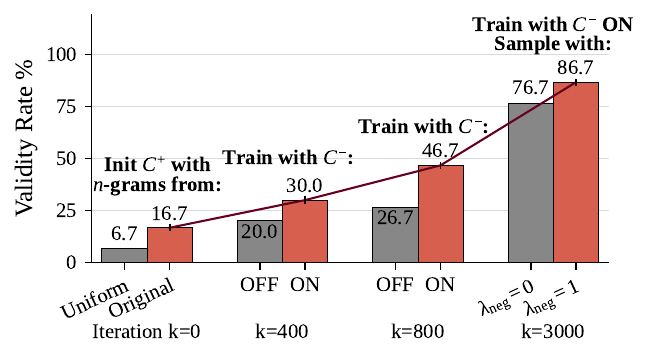}
    \includegraphics[width=0.48\textwidth]{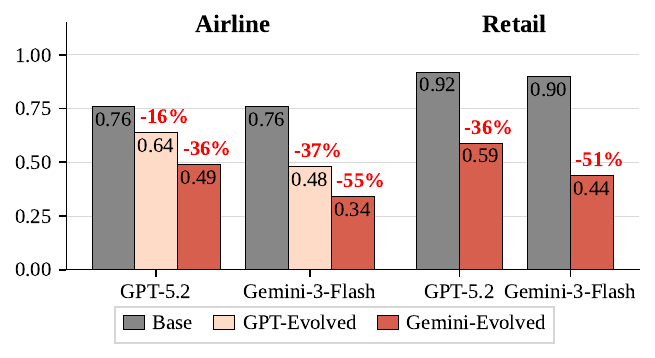}
    \caption{\textbf{$n$-gram Model (Left):} Acceptance rate of sampled sequences across training stages and model configurations on the Airline domain. The dark red line traces the full adaptive contrastive model across iterations. \textbf{Task Evolution (Right):} Success rate of two agents (Gemini-3-Flash as the user simulator) on base vs.\ evolved tasks. Tasks are evolved by either GPT-5.2 or Gemini-3.}
    \label{fig:method_analysis}
\vspace{-1em}
\end{figure*}

\paragraph{Ablation of the Adaptive Contrastive $n$-gram Model.}
The goal of using $n$-gram model in TASTE is to sample from the space of valid tool sequences while maintaining broad coverage. We achieved $86.7\%$ validity rate with our full trained model compared to uniform tool sampling ($6.7\%$), illustrates the overall impact of our design.
We extend the standard model in two ways: through adaptive online training and the integration of contrastive negative evidence. Figure~\ref{fig:method_analysis} (left) presents an ablation study of the model's components within the airline domain. Adaptive online training is essential: even when the model is initialized with $n$-grams from \tbv tasks, the validity rate without training reaches only $16.7\%$, whereas iterative training steadily improves it (as shown by the dark red line). The contrastive formulation provides a complementary and substantial gain. Training with negative evidence (i.e., ingesting invalid $n$-grams into $C^-$ to avoid sampling them) improves the validity rate by $10$ points at $k\!=\!400$, and by $20$ points at $k=800$. Finally, evaluating the final model ($k\!=\!3000$) trained with negative evidence to examine the impact of $\lambda_{\text{neg}}$ on sampling reveals an additional $10$-point improvement. In total, our design choices raise the validity rate from $6.7\%$ to $86.7\%$. This ${\sim}\!13{\times}$ improvement enables us to sample a large pool of plausibly valid candidates.

\paragraph{Evaluation of the Verifier Agent.} To validate generated tasks, we apply a hint-assisted verifier agent, which attempts to complete each task given a shuffled, partially masked list of gold tool calls (Section~\ref{sub:task_generation}). A task is deemed valid if the agent succeeds. To evaluate its reliability, we construct a validation dataset by comparing each original task from \taubench with its corresponding entry in \taubenchverified, and assign these labels: \emph{valid:} if the task was not modified in Verified; \emph{invalid:} if the fix involved changing an action or substantially rewriting the instruction; tasks with minor fixes (e.g., capitalizing a single word) are excluded. This yields $50$ Airline tasks ($8$ invalid) and $86$ Retail tasks ($9$ invalid). The verifier agent achieves precision of $1.0$ and $0.97$ and recall of $0.75$ and $0.83$ in Airline and Retail, respectively. Accordingly, the agent may reject some valid tasks (recall $<0.9$), but the near-perfect precision ensures any task passing is almost certainly valid. Finally, we also manually examined all tasks from \ourbenchmark on which none of the evaluated agent-user pairs succeeded ($1$ Airline, $7$ Retail, and $7$ Telecom). All $15$ tasks were found to be valid; failures were due to agent mistakes.

 \paragraph{Direct LLM Sequence Baseline.} We asked GPT-6 Luna to generate $100$ Airline tool sequences and compared them with $100$ TASTE samples of the same lengths. TASTE produced $68$ distinct sequences, compared with $56$ from direct prompting. TASTE also had higher TTR ($0.84$ versus $0.49$), with mean WED of $3.51$ versus $3.34$. We did not assess plausibility or generate complete tasks in this comparison (Appendix~\ref{app:direct-llm}).

 \paragraph{Creating Difficult Tasks.} TASTE increases task difficulty through task evolution. In Figure~\ref{fig:method_analysis} (right), we compare base tasks in the Airline and Retail domains with evolved versions, using two agents and the Gemini-$3$-Flash user simulator. As shown, success rates are $36$--$55$\% lower when tasks are evolved by Gemini-$3$, and $16$--$37$\% lower when evolved by GPT-$5.2$, leading us to select Gemini-$3$.  Our model-choice analysis concerns task evolution; sensitivity to the base-task generator and verifier remains to be established.  Difficulty can also be controlled through tool-sequence length and the read-to-write tool ratio during TASTE's sampling and selection stages. In Figure~\ref{fig:difficulty-desiderata} (Appendix~\ref{app:difficulty-desiderata}), we compare two tiers of tasks (top $50\%$ vs. bottom $50\%$ by length and tool type ratio). As shown, performance on longer tasks is $30\%$ lower on average (across all agent-user pairs and domains) and $43\%$ lower on write-heavy tasks. This suggests that TASTE can be used to generate even harder tasks in the future, should existing benchmarks saturate.  Base/evolved examples from all three domains are provided in Appendix~\ref{app:examples}.

\subsection{Extension to AppWorld}
\label{sub:appworld}

We apply TASTE to the complementary benchmark AppWorld, where autonomous agents write code over application APIs. Sequence sampling and clustering are retained, while task realization is adapted to AppWorld's evaluation framework. This demonstrates TASTE's applicability beyond conversational tool use. Similar to the \taubench results, performance on the hardened tasks is consistently and substantially lower than on the base tasks, with relative drops of $28.6$--$40.5\%$ across the four evaluated agents.

These preliminary results are summarized in Table~\ref{tab:appworld-results}; evaluation details and limitations are discussed in Appendix~\ref{app:appworld}.

\section{Related Work}

Tool-use agent benchmarks vary along several axes. Early benchmarks evaluate single-turn tool selection and argument grounding over API collections \citep{toolllm24qin, bfcl_25_patil, apibank23Li, ToolQA23Zhuang, mcpverse25lei}. Recent work involves multi-turn interactions in persistent environments, such as web navigation and code repositories \citep{webarena_24_zhou, swebench_24_Jimenez}.
Our focus is on conversational benchmarks that pair an agent with a simulated user over a mutable environment, and evaluate the resulting world state after the full interaction \citep{taubench_24_yao, tau2barres, toolsandbox25lu}. While valuable, existing agent benchmarks rely heavily on manually authored tasks, a process that is costly, error-prone, and difficult to scale as agent capabilities advance \citep{saber2025cuadron, bestpractices2025Zhu, yehudai2026agenticclearautomatingmultilevel}.

Synthetic generation methods reduce annotation cost. Early approaches bootstrap instructions from seed sets \citep{selfinstruct23wang, unnaturalinstructions23honovich, wizardlm24xu}, while tool-use extensions generate gold tool calls through insertion filtering \citep{toolformer23shick}, synthesis from tool descriptions \citep{toolllm24qin, toolalpaca23tang}, and post-hoc verification \citep{apigen24liu}. Recent pipelines scale to multi-turn using task blueprints, simulated environments, multi-agent role-playing, and trajectory feedback \citep{apigenmt25prabhakar, toucan25xu, closetheloop25li, implicitexperience26xu, toolforge25chen, LAM25Hoang}. Closest to our approach are ToolGrad \citep{toolgrad25zhou} and Trajectory2Task \citep{trajectory2task}, which synthesize tasks from trajectories. ToolGrad iteratively builds tool chains by proposing, executing, and selecting the best API call at each step, extending the sequence until completion before instantiating the user instruction. Trajectory2Task starts from an existing trajectory, progressively replaces tool calls, and evolves user instructions toward challenging intents. Both primarily employ reverse synthesis to increase the plausibility of generating a valid task and to produce training data. Our motivation is different: we start from tool sequences to explicitly explore and cover the space of tool combinations, and our focus is on constructing better agent benchmarks rather than training data.

We define coverage in terms of gold tool sequences, adopting diversity measures from the NLP literature \citep{raven2023mccoy, evaluatingnovelty24merrill, standardizingdiversity24shaib, evaluatingdiversity21tevet} over tools rather than words. Existing benchmarks typically summarize diversity by domain, tool, or schema counts, and characterize difficulty through capability tags \citep{toolllm24qin, mcpverse25lei, toolsandbox25lu, tripbench26Shen}. Such summaries indicate which tools are available, but not whether tasks exercise diverse procedural patterns of tool use. In contrast, sequence-level coverage allows us to measure whether a benchmark explores a broad range of agent behaviors rather than repeatedly testing similar patterns.

\section{Conclusions}

In this work, we proposed TASTE, an automatic method for generating agent benchmark tasks that are difficult and cover a broad range of tool-use patterns. Using TASTE, we constructed \ourbenchmark, a challenging extension of \taubench, and showed that models nearly saturating the original benchmark suffer performance drops on our generated tasks. Our focus was on conversational tool-using agents. Despite this focus, TASTE operates on tool sequences and environment specifications, making it readily adaptable to single-turn and non-conversational settings. Future work could apply TASTE to generate training data in addition to evaluation benchmarks, and extend the coverage and difficulty definitions to additional axes beyond those explored here.

\begin{ack}
This research was conducted in collaboration with the Technion and IBM Research. The work was supported by the IBM-Technion Research collaboration.
\end{ack}

\bibliographystyle{unsrtnat}
\bibliography{cite.bib}

@inproceedings{appworld24trivedi,
  title        = {{AppWorld}: A Controllable World of Apps and People for Benchmarking Interactive Coding Agents},
  author       = {Harsh Trivedi and Tushar Khot and Mareike Hartmann and Ruskin Manku and Vinty Dong and Edward Li and Shashank Gupta and Ashish Sabharwal and Niranjan Balasubramanian},
  booktitle    = {Proceedings of the 62nd Annual Meeting of the Association for Computational Linguistics (Volume 1: Long Papers)},
  year         = {2024},
  pages        = {16022--16076},
  publisher    = {Association for Computational Linguistics},
  url          = {https://aclanthology.org/2024.acl-long.850/},
  doi          = {10.18653/v1/2024.acl-long.850}
}

@inproceedings{swebench_24_Jimenez,
  author       = {Carlos E. Jimenez and
                  John Yang and
                  Alexander Wettig and
                  Shunyu Yao and
                  Kexin Pei and
                  Ofir Press and
                  Karthik R. Narasimhan},
  title        = {SWE-bench: Can Language Models Resolve Real-world Github Issues?},
  booktitle    = {The Twelfth International Conference on Learning Representations,
                  {ICLR} 2024, Vienna, Austria, May 7-11, 2024},
  publisher    = {OpenReview.net},
  year         = {2024},
  url          = {https://openreview.net/forum?id=VTF8yNQM66},
  bibsource    = {dblp computer science bibliography, https://dblp.org}
}

@article{trajectory2task,
  author       = {Ziyi Wang and
                  Yuxuan Lu and
                  Yimeng Zhang and
                  Ziwei Dong and
                  Jing Huang and
                  Jiri Gesi and
                  Xianfeng Tang and
                  Chen Luo and
                  Yisi Sang and
                  Hanqing Lu and
                  Manling Li and
                  Dakuo Wang},
  title        = {Trajectory2Task: Training Robust Tool-Calling Agents with Synthesized
                  Yet Verifiable Data for Complex User Intents},
  journal      = {CoRR},
  volume       = {abs/2601.20144},
  year         = {2026},
  url          = {https://doi.org/10.48550/arXiv.2601.20144},
  doi          = {10.48550/ARXIV.2601.20144},
  eprinttype   = {arXiv},
  eprint       = {2601.20144},
  bibsource    = {dblp computer science bibliography, https://dblp.org}
}

@article{raven2023mccoy,
  author       = {R. Thomas McCoy and
                  Paul Smolensky and
                  Tal Linzen and
                  Jianfeng Gao and
                  Asli Celikyilmaz},
  title        = {How Much Do Language Models Copy From Their Training Data? Evaluating
                  Linguistic Novelty in Text Generation Using {RAVEN}},
  journal      = {Trans. Assoc. Comput. Linguistics},
  volume       = {11},
  pages        = {652--670},
  year         = {2023},
  url          = {https://doi.org/10.1162/tacl\_a\_00567},
  doi          = {10.1162/TACL\_A\_00567},
  bibsource    = {dblp computer science bibliography, https://dblp.org}
}

@article{saber2025cuadron,
  author       = {Alejandro Cuadron and
                  Pengfei Yu and
                  Yang Liu and
                  Arpit Gupta},
  title        = {{SABER:} Small Actions, Big Errors - Safeguarding Mutating Steps in
                  {LLM} Agents},
  journal      = {CoRR},
  volume       = {abs/2512.07850},
  year         = {2025},
  url          = {https://doi.org/10.48550/arXiv.2512.07850},
  doi          = {10.48550/ARXIV.2512.07850},
  eprinttype   = {arXiv},
  eprint       = {2512.07850},
  bibsource    = {dblp computer science bibliography, https://dblp.org}
}

@article{taubench_24_yao,
  author       = {Shunyu Yao and
                  Noah Shinn and
                  Pedram Razavi and
                  Karthik Narasimhan},
  title        = {{\(\tau\)}-bench: {A} Benchmark for Tool-Agent-User Interaction in
                  Real-World Domains},
  journal      = {CoRR},
  volume       = {abs/2406.12045},
  year         = {2024},
  url          = {https://doi.org/10.48550/arXiv.2406.12045},
  doi          = {10.48550/ARXIV.2406.12045},
  eprinttype   = {arXiv},
  eprint       = {2406.12045},
  bibsource    = {dblp computer science bibliography, https://dblp.org}
}

@inproceedings{webarena_24_zhou,
  author       = {Shuyan Zhou and
                  Frank F. Xu and
                  Hao Zhu and
                  Xuhui Zhou and
                  Robert Lo and
                  Abishek Sridhar and
                  Xianyi Cheng and
                  Tianyue Ou and
                  Yonatan Bisk and
                  Daniel Fried and
                  Uri Alon and
                  Graham Neubig},
  title        = {WebArena: {A} Realistic Web Environment for Building Autonomous Agents},
  booktitle    = {The Twelfth International Conference on Learning Representations,
                  {ICLR} 2024, Vienna, Austria, May 7-11, 2024},
  publisher    = {OpenReview.net},
  year         = {2024},
  url          = {https://openreview.net/forum?id=oKn9c6ytLx},
  bibsource    = {dblp computer science bibliography, https://dblp.org}
}

@article{kmedoids,
  author       = {Hae{-}Sang Park and
                  Chi{-}Hyuck Jun},
  title        = {A simple and fast algorithm for K-medoids clustering},
  journal      = {Expert Syst. Appl.},
  volume       = {36},
  number       = {2},
  pages        = {3336--3341},
  year         = {2009},
  url          = {https://doi.org/10.1016/j.eswa.2008.01.039},
  doi          = {10.1016/J.ESWA.2008.01.039},
  bibsource    = {dblp computer science bibliography, https://dblp.org}
}

@article{tau2barres,
  author       = {Victor Barres and
                  Honghua Dong and
                  Soham Ray and
                  Xujie Si and
                  Karthik Narasimhan},
  title        = {{\(\tau\)}\({}^{\mbox{2}}\)-Bench: Evaluating Conversational Agents
                  in a Dual-Control Environment},
  journal      = {CoRR},
  volume       = {abs/2506.07982},
  year         = {2025},
  url          = {https://doi.org/10.48550/arXiv.2506.07982},
  doi          = {10.48550/ARXIV.2506.07982},
  eprinttype   = {arXiv},
  eprint       = {2506.07982},
  bibsource    = {dblp computer science bibliography, https://dblp.org}
}

@article{yehudai2025surveyevaluationllmbasedagents,
  author       = {Asaf Yehudai and
                  Lilach Eden and
                  Alan Li and
                  Guy Uziel and
                  Yilun Zhao and
                  Roy Bar{-}Haim and
                  Arman Cohan and
                  Michal Shmueli{-}Scheuer},
  title        = {Survey on Evaluation of LLM-based Agents},
  journal      = {CoRR},
  volume       = {abs/2503.16416},
  year         = {2025},
  url          = {https://doi.org/10.48550/arXiv.2503.16416},
  doi          = {10.48550/ARXIV.2503.16416},
  eprinttype   = {arXiv},
  eprint       = {2503.16416},
  bibsource    = {dblp computer science bibliography, https://dblp.org}
}

@article{pam_kaufman1990,
  title={Finding groups in data},
  author={Rousseeuw, Peter J and Kaufman, L},
  journal={Hoboken: Wiley Online Library},
  volume={1},
  pages={371},
  year={1990},
  publisher={Wiley Online Library}
}

@article{bestpractices2025Zhu,
  author       = {Yuxuan Zhu and
                  Tengjun Jin and
                  Yada Pruksachatkun and
                  Andy Zhang and
                  Shu Liu and
                  Sasha Cui and
                  Sayash Kapoor and
                  Shayne Longpre and
                  Kevin Meng and
                  Rebecca Weiss and
                  Fazl Barez and
                  Rahul Gupta and
                  Jwala Dhamala and
                  Jacob Merizian and
                  Mario Giulianelli and
                  Harry Coppock and
                  Cozmin Ududec and
                  Jasjeet S. Sekhon and
                  Jacob Steinhardt and
                  Antony Kellerman and
                  Sarah Schwettmann and
                  Matei Zaharia and
                  Ion Stoica and
                  Percy Liang and
                  Daniel Kang},
  title        = {Establishing Best Practices for Building Rigorous Agentic Benchmarks},
  journal      = {CoRR},
  volume       = {abs/2507.02825},
  year         = {2025},
  url          = {https://doi.org/10.48550/arXiv.2507.02825},
  doi          = {10.48550/ARXIV.2507.02825},
  eprinttype   = {arXiv},
  eprint       = {2507.02825},
  bibsource    = {dblp computer science bibliography, https://dblp.org}
}

@misc{yehudai2026agenticclearautomatingmultilevel,
      title={Agentic CLEAR: Automating Multi-Level Evaluation of LLM Agents},
      author={Asaf Yehudai and Lilach Eden and Michal Shmueli-Scheuer},
      year={2026},
      eprint={2605.22608},
      archivePrefix={arXiv},
      primaryClass={cs.CL},
      url={https://arxiv.org/abs/2605.22608},
}

@inproceedings{Mohammadi_survery_2025,
  author       = {Mahmoud Mohammadi and
                  Yipeng Li and
                  Jane Lo and
                  Wendy Yip},
  editor       = {Luiza Antonie and
                  Jian Pei and
                  Xiaohui Yu and
                  Flavio Chierichetti and
                  Hady W. Lauw and
                  Yizhou Sun and
                  Srinivasan Parthasarathy},
  title        = {Evaluation and Benchmarking of {LLM} Agents: {A} Survey},
  booktitle    = {Proceedings of the 31st {ACM} {SIGKDD} Conference on Knowledge Discovery
                  and Data Mining, V.2, {KDD} 2025, Toronto ON, Canada, August 3-7,
                  2025},
  pages        = {6129--6139},
  publisher    = {{ACM}},
  year         = {2025},
  url          = {https://doi.org/10.1145/3711896.3736570},
  doi          = {10.1145/3711896.3736570},
  bibsource    = {dblp computer science bibliography, https://dblp.org}
}

@inproceedings{apibank23Li,
  author       = {Minghao Li and
                  Yingxiu Zhao and
                  Bowen Yu and
                  Feifan Song and
                  Hangyu Li and
                  Haiyang Yu and
                  Zhoujun Li and
                  Fei Huang and
                  Yongbin Li},
  editor       = {Houda Bouamor and
                  Juan Pino and
                  Kalika Bali},
  title        = {API-Bank: {A} Comprehensive Benchmark for Tool-Augmented LLMs},
  booktitle    = {Proceedings of the 2023 Conference on Empirical Methods in Natural
                  Language Processing, {EMNLP} 2023, Singapore, December 6-10, 2023},
  pages        = {3102--3116},
  publisher    = {Association for Computational Linguistics},
  year         = {2023},
  url          = {https://doi.org/10.18653/v1/2023.emnlp-main.187},
  doi          = {10.18653/V1/2023.EMNLP-MAIN.187},
  bibsource    = {dblp computer science bibliography, https://dblp.org}
}

@inproceedings{ToolQA23Zhuang,
  author       = {Yuchen Zhuang and
                  Yue Yu and
                  Kuan Wang and
                  Haotian Sun and
                  Chao Zhang},
  editor       = {Alice Oh and
                  Tristan Naumann and
                  Amir Globerson and
                  Kate Saenko and
                  Moritz Hardt and
                  Sergey Levine},
  title        = {ToolQA: {A} Dataset for {LLM} Question Answering with External Tools},
  booktitle    = {Advances in Neural Information Processing Systems 36: Annual Conference
                  on Neural Information Processing Systems 2023, NeurIPS 2023, New Orleans,
                  LA, USA, December 10 - 16, 2023},
  year         = {2023},
  url          = {http://papers.nips.cc/paper\_files/paper/2023/hash/9cb2a7495900f8b602cb10159246a016-Abstract-Datasets\_and\_Benchmarks.html},
  bibsource    = {dblp computer science bibliography, https://dblp.org}
}

@inproceedings{bfcl_25_patil,
  author       = {Shishir G. Patil and
                  Huanzhi Mao and
                  Fanjia Yan and
                  Charlie Cheng{-}Jie Ji and
                  Vishnu Suresh and
                  Ion Stoica and
                  Joseph E. Gonzalez},
  editor       = {Aarti Singh and
                  Maryam Fazel and
                  Daniel Hsu and
                  Simon Lacoste{-}Julien and
                  Felix Berkenkamp and
                  Tegan Maharaj and
                  Kiri Wagstaff and
                  Jerry Zhu},
  title        = {The Berkeley Function Calling Leaderboard {(BFCL):} From Tool Use
                  to Agentic Evaluation of Large Language Models},
  booktitle    = {Forty-second International Conference on Machine Learning, {ICML}
                  2025, Vancouver, BC, Canada, July 13-19, 2025},
  series       = {Proceedings of Machine Learning Research},
  publisher    = {{PMLR} / OpenReview.net},
  year         = {2025},
  url          = {https://proceedings.mlr.press/v267/patil25a.html},
  bibsource    = {dblp computer science bibliography, https://dblp.org}
}

@inproceedings{toolsandbox25lu,
  author       = {Jiarui Lu and
                  Thomas Holleis and
                  Yizhe Zhang and
                  Bernhard Aumayer and
                  Feng Nan and
                  Haoping Bai and
                  Shuang Ma and
                  Shen Ma and
                  Mengyu Li and
                  Guoli Yin and
                  Zirui Wang and
                  Ruoming Pang},
  editor       = {Luis Chiruzzo and
                  Alan Ritter and
                  Lu Wang},
  title        = {ToolSandbox: {A} Stateful, Conversational, Interactive Evaluation
                  Benchmark for {LLM} Tool Use Capabilities},
  booktitle    = {Findings of the Association for Computational Linguistics: {NAACL}
                  2025, Albuquerque, New Mexico, USA, April 29 - May 4, 2025},
  series       = {Findings of {ACL}},
  pages        = {1160--1183},
  publisher    = {Association for Computational Linguistics},
  year         = {2025},
  url          = {https://doi.org/10.18653/v1/2025.findings-naacl.65},
  doi          = {10.18653/V1/2025.FINDINGS-NAACL.65},
  bibsource    = {dblp computer science bibliography, https://dblp.org}
}

@article{mcpverse25lei,
  author       = {Fei Lei and
                  Yibo Yang and
                  Wenxiu Sun and
                  Dahua Lin},
  title        = {MCPVerse: An Expansive, Real-World Benchmark for Agentic Tool Use},
  journal      = {CoRR},
  volume       = {abs/2508.16260},
  year         = {2025},
  url          = {https://doi.org/10.48550/arXiv.2508.16260},
  doi          = {10.48550/ARXIV.2508.16260},
  eprinttype   = {arXiv},
  eprint       = {2508.16260},
  bibsource    = {dblp computer science bibliography, https://dblp.org}
}

@article{tripbench26Shen,
  author       = {Yuanzhe Shen and
                  Zisu Huang and
                  Zhengyuan Wang and
                  Muzhao Tian and
                  Zhengkang Guo and
                  Chenyang Zhang and
                  Shuaiyu Zhou and
                  Zengjie Hu and
                  Dailin Li and
                  Jingwen Xu and
                  Kaimin Wang and
                  Wenhao Liu and
                  Tianlong Li and
                  Fengpeng Yue and
                  Feng Hong and
                  Cao Liu and
                  Ke Zeng},
  title        = {TRIP-Bench: {A} Benchmark for Long-Horizon Interactive Agents in Real-World
                  Scenarios},
  journal      = {CoRR},
  volume       = {abs/2602.01675},
  year         = {2026},
  url          = {https://doi.org/10.48550/arXiv.2602.01675},
  doi          = {10.48550/ARXIV.2602.01675},
  eprinttype   = {arXiv},
  eprint       = {2602.01675},
  bibsource    = {dblp computer science bibliography, https://dblp.org}
}

@inproceedings{selfinstruct23wang,
  author       = {Yizhong Wang and
                  Yeganeh Kordi and
                  Swaroop Mishra and
                  Alisa Liu and
                  Noah A. Smith and
                  Daniel Khashabi and
                  Hannaneh Hajishirzi},
  editor       = {Anna Rogers and
                  Jordan L. Boyd{-}Graber and
                  Naoaki Okazaki},
  title        = {Self-Instruct: Aligning Language Models with Self-Generated Instructions},
  booktitle    = {Proceedings of the 61st Annual Meeting of the Association for Computational
                  Linguistics (Volume 1: Long Papers), {ACL} 2023, Toronto, Canada,
                  July 9-14, 2023},
  pages        = {13484--13508},
  publisher    = {Association for Computational Linguistics},
  year         = {2023},
  url          = {https://doi.org/10.18653/v1/2023.acl-long.754},
  doi          = {10.18653/V1/2023.ACL-LONG.754},
  bibsource    = {dblp computer science bibliography, https://dblp.org}
}

@article{toolgrad25zhou,
  author       = {Zhongyi Zhou and
                  Kohei Uehara and
                  Haoyu Zhang and
                  Jingtao Zhou and
                  Lin Gu and
                  Ruofei Du and
                  Zheng Xu and
                  Tatsuya Harada},
  title        = {ToolGrad: Efficient Tool-use Dataset Generation with Textual "Gradients"},
  journal      = {CoRR},
  volume       = {abs/2508.04086},
  year         = {2025},
  url          = {https://doi.org/10.48550/arXiv.2508.04086},
  doi          = {10.48550/ARXIV.2508.04086},
  eprinttype   = {arXiv},
  eprint       = {2508.04086},
  bibsource    = {dblp computer science bibliography, https://dblp.org}
}

@inproceedings{LAM25Hoang,
  author       = {Thai Quoc Hoang and
                  Kung{-}Hsiang Huang and
                  Shirley Kokane and
                  Jianguo Zhang and
                  Zuxin Liu and
                  Ming Zhu and
                  Jake Grigsby and
                  Tian Lan and
                  Michael S. Ryoo and
                  Chien{-}Sheng Wu and
                  Shelby Heinecke and
                  Huan Wang and
                  Silvio Savarese and
                  Caiming Xiong and
                  Juan Carlos Niebles},
  editor       = {Wanxiang Che and
                  Joyce Nabende and
                  Ekaterina Shutova and
                  Mohammad Taher Pilehvar},
  title        = {{LAM} {SIMULATOR:} Advancing Data Generation for Large Action Model
                  Training via Online Exploration and Trajectory Feedback},
  booktitle    = {Findings of the Association for Computational Linguistics, {ACL} 2025,
                  Vienna, Austria, July 27 - August 1, 2025},
  series       = {Findings of {ACL}},
  pages        = {12921--12934},
  publisher    = {Association for Computational Linguistics},
  year         = {2025},
  url          = {https://aclanthology.org/2025.findings-acl.670/},
  bibsource    = {dblp computer science bibliography, https://dblp.org}
}

@article{toolforge25chen,
  author       = {Hao Chen and
                  Zhexin Hu and
                  Jiajun Chai and
                  Haocheng Yang and
                  Hang He and
                  Xiaohan Wang and
                  Wei Lin and
                  Luhang Wang and
                  Guojun Yin and
                  Zhuofeng Zhao},
  title        = {ToolForge: {A} Data Synthesis Pipeline for Multi-Hop Search without
                  Real-World APIs},
  journal      = {CoRR},
  volume       = {abs/2512.16149},
  year         = {2025},
  url          = {https://doi.org/10.48550/arXiv.2512.16149},
  doi          = {10.48550/ARXIV.2512.16149},
  eprinttype   = {arXiv},
  eprint       = {2512.16149},
  bibsource    = {dblp computer science bibliography, https://dblp.org}
}

@article{toucan25xu,
  author       = {Zhangchen Xu and
                  Adriana Meza Soria and
                  Shawn Tan and
                  Anurag Roy and
                  Ashish Sunil Agrawal and
                  Radha Poovendran and
                  Rameswar Panda},
  title        = {{TOUCAN:} Synthesizing 1.5M Tool-Agentic Data from Real-World {MCP}
                  Environments},
  journal      = {CoRR},
  volume       = {abs/2510.01179},
  year         = {2025},
  url          = {https://doi.org/10.48550/arXiv.2510.01179},
  doi          = {10.48550/ARXIV.2510.01179},
  eprinttype   = {arXiv},
  eprint       = {2510.01179},
  bibsource    = {dblp computer science bibliography, https://dblp.org}
}

@article{closetheloop25li,
  author       = {Yuwen Li and
                  Wei Zhang and
                  Zelong Huang and
                  Mason Yang and
                  Jiajun Wu and
                  Shawn Guo and
                  Huahao Hu and
                  Lingyi Sun and
                  Jian Yang and
                  Mingjie Tang and
                  Byran Dai},
  title        = {Close the Loop: Synthesizing Infinite Tool-Use Data via Multi-Agent
                  Role-Playing},
  journal      = {CoRR},
  volume       = {abs/2512.23611},
  year         = {2025},
  url          = {https://doi.org/10.48550/arXiv.2512.23611},
  doi          = {10.48550/ARXIV.2512.23611},
  eprinttype   = {arXiv},
  eprint       = {2512.23611},
  bibsource    = {dblp computer science bibliography, https://dblp.org}
}

@inproceedings{wizardlm24xu,
  author       = {Can Xu and
                  Qingfeng Sun and
                  Kai Zheng and
                  Xiubo Geng and
                  Pu Zhao and
                  Jiazhan Feng and
                  Chongyang Tao and
                  Qingwei Lin and
                  Daxin Jiang},
  title        = {WizardLM: Empowering Large Pre-Trained Language Models to Follow Complex
                  Instructions},
  booktitle    = {The Twelfth International Conference on Learning Representations,
                  {ICLR} 2024, Vienna, Austria, May 7-11, 2024},
  publisher    = {OpenReview.net},
  year         = {2024},
  url          = {https://openreview.net/forum?id=CfXh93NDgH},
  bibsource    = {dblp computer science bibliography, https://dblp.org}
}

@article{apigenmt25prabhakar,
  author       = {Akshara Prabhakar and
                  Zuxin Liu and
                  Ming Zhu and
                  Jianguo Zhang and
                  Tulika Awalgaonkar and
                  Shiyu Wang and
                  Zhiwei Liu and
                  Haolin Chen and
                  Thai Hoang and
                  Juan Carlos Niebles and
                  Shelby Heinecke and
                  Weiran Yao and
                  Huan Wang and
                  Silvio Savarese and
                  Caiming Xiong},
  title        = {APIGen-MT: Agentic Pipeline for Multi-Turn Data Generation via Simulated
                  Agent-Human Interplay},
  journal      = {CoRR},
  volume       = {abs/2504.03601},
  year         = {2025},
  url          = {https://doi.org/10.48550/arXiv.2504.03601},
  doi          = {10.48550/ARXIV.2504.03601},
  eprinttype   = {arXiv},
  eprint       = {2504.03601},
  bibsource    = {dblp computer science bibliography, https://dblp.org}
}

@inproceedings{toolformer23shick,
  author       = {Timo Schick and
                  Jane Dwivedi{-}Yu and
                  Roberto Dess{\`{\i}} and
                  Roberta Raileanu and
                  Maria Lomeli and
                  Eric Hambro and
                  Luke Zettlemoyer and
                  Nicola Cancedda and
                  Thomas Scialom},
  editor       = {Alice Oh and
                  Tristan Naumann and
                  Amir Globerson and
                  Kate Saenko and
                  Moritz Hardt and
                  Sergey Levine},
  title        = {Toolformer: Language Models Can Teach Themselves to Use Tools},
  booktitle    = {Advances in Neural Information Processing Systems 36: Annual Conference
                  on Neural Information Processing Systems 2023, NeurIPS 2023, New Orleans,
                  LA, USA, December 10 - 16, 2023},
  year         = {2023},
  url          = {http://papers.nips.cc/paper\_files/paper/2023/hash/d842425e4bf79ba039352da0f658a906-Abstract-Conference.html},
  bibsource    = {dblp computer science bibliography, https://dblp.org}
}

@article{toolalpaca23tang,
  author       = {Qiaoyu Tang and
                  Ziliang Deng and
                  Hongyu Lin and
                  Xianpei Han and
                  Qiao Liang and
                  Le Sun},
  title        = {ToolAlpaca: Generalized Tool Learning for Language Models with 3000
                  Simulated Cases},
  journal      = {CoRR},
  volume       = {abs/2306.05301},
  year         = {2023},
  url          = {https://doi.org/10.48550/arXiv.2306.05301},
  doi          = {10.48550/ARXIV.2306.05301},
  eprinttype   = {arXiv},
  eprint       = {2306.05301},
  bibsource    = {dblp computer science bibliography, https://dblp.org}
}

@inproceedings{toolllm24qin,
  author       = {Yujia Qin and
                  Shihao Liang and
                  Yining Ye and
                  Kunlun Zhu and
                  Lan Yan and
                  Yaxi Lu and
                  Yankai Lin and
                  Xin Cong and
                  Xiangru Tang and
                  Bill Qian and
                  Sihan Zhao and
                  Lauren Hong and
                  Runchu Tian and
                  Ruobing Xie and
                  Jie Zhou and
                  Mark Gerstein and
                  Dahai Li and
                  Zhiyuan Liu and
                  Maosong Sun},
  title        = {ToolLLM: Facilitating Large Language Models to Master 16000+ Real-world
                  APIs},
  booktitle    = {The Twelfth International Conference on Learning Representations,
                  {ICLR} 2024, Vienna, Austria, May 7-11, 2024},
  publisher    = {OpenReview.net},
  year         = {2024},
  url          = {https://openreview.net/forum?id=dHng2O0Jjr},
  bibsource    = {dblp computer science bibliography, https://dblp.org}
}

@inproceedings{apigen24liu,
  author       = {Zuxin Liu and
                  Thai Hoang and
                  Jianguo Zhang and
                  Ming Zhu and
                  Tian Lan and
                  Shirley Kokane and
                  Juntao Tan and
                  Weiran Yao and
                  Zhiwei Liu and
                  Yihao Feng and
                  Rithesh R. N. and
                  Liangwei Yang and
                  Silvio Savarese and
                  Juan Carlos Niebles and
                  Huan Wang and
                  Shelby Heinecke and
                  Caiming Xiong},
  editor       = {Amir Globersons and
                  Lester Mackey and
                  Danielle Belgrave and
                  Angela Fan and
                  Ulrich Paquet and
                  Jakub M. Tomczak and
                  Cheng Zhang},
  title        = {APIGen: Automated PIpeline for Generating Verifiable and Diverse Function-Calling
                  Datasets},
  booktitle    = {Advances in Neural Information Processing Systems 38: Annual Conference
                  on Neural Information Processing Systems 2024, NeurIPS 2024, Vancouver,
                  BC, Canada, December 10 - 15, 2024},
  year         = {2024},
  url          = {http://papers.nips.cc/paper\_files/paper/2024/hash/61cce86d180b1184949e58939c4f983d-Abstract-Datasets\_and\_Benchmarks\_Track.html},
  bibsource    = {dblp computer science bibliography, https://dblp.org}
}

@article{implicitexperience26xu,
  author       = {Zhihao Xu and
                  Rumei Li and
                  Jiahuan Li and
                  Rongxiang Weng and
                  Jingang Wang and
                  Xunliang Cai and
                  Xiting Wang},
  title        = {Unlocking Implicit Experience: Synthesizing Tool-Use Trajectories
                  from Text},
  journal      = {CoRR},
  volume       = {abs/2601.10355},
  year         = {2026},
  url          = {https://doi.org/10.48550/arXiv.2601.10355},
  doi          = {10.48550/ARXIV.2601.10355},
  eprinttype   = {arXiv},
  eprint       = {2601.10355},
  bibsource    = {dblp computer science bibliography, https://dblp.org}
}

@inproceedings{unnaturalinstructions23honovich,
  author       = {Or Honovich and
                  Thomas Scialom and
                  Omer Levy and
                  Timo Schick},
  editor       = {Anna Rogers and
                  Jordan L. Boyd{-}Graber and
                  Naoaki Okazaki},
  title        = {Unnatural Instructions: Tuning Language Models with (Almost) No Human
                  Labor},
  booktitle    = {Proceedings of the 61st Annual Meeting of the Association for Computational
                  Linguistics (Volume 1: Long Papers), {ACL} 2023, Toronto, Canada,
                  July 9-14, 2023},
  pages        = {14409--14428},
  publisher    = {Association for Computational Linguistics},
  year         = {2023},
  url          = {https://doi.org/10.18653/v1/2023.acl-long.806},
  doi          = {10.18653/V1/2023.ACL-LONG.806},
  bibsource    = {dblp computer science bibliography, https://dblp.org}
}

@article{lmars25wang,
  author       = {Ziqi Wang and
                  Boqin Yuan},
  title        = {{L-MARS:} Legal Multi-Agent Workflow with Orchestrated Reasoning and
                  Agentic Search},
  journal      = {CoRR},
  volume       = {abs/2509.00761},
  year         = {2025},
  url          = {https://doi.org/10.48550/arXiv.2509.00761},
  doi          = {10.48550/ARXIV.2509.00761},
  eprinttype   = {arXiv},
  eprint       = {2509.00761},
  bibsource    = {dblp computer science bibliography, https://dblp.org}
}

@article{skyrl25cao,
  author       = {Shiyi Cao and
                  Dacheng Li and
                  Fangzhou Zhao and
                  Shuo Yuan and
                  Sumanth Hegde and
                  Connor Chen and
                  Charlie Ruan and
                  Tyler Griggs and
                  Shu Liu and
                  Eric Tang and
                  Richard Liaw and
                  Philipp Moritz and
                  Matei Zaharia and
                  Joseph E. Gonzalez and
                  Ion Stoica},
  title        = {SkyRL-Agent: Efficient {RL} Training for Multi-turn {LLM} Agent},
  journal      = {CoRR},
  volume       = {abs/2511.16108},
  year         = {2025},
  url          = {https://doi.org/10.48550/arXiv.2511.16108},
  doi          = {10.48550/ARXIV.2511.16108},
  eprinttype   = {arXiv},
  eprint       = {2511.16108},
  bibsource    = {dblp computer science bibliography, https://dblp.org}
}

@inproceedings{sweagent24yang,
  author       = {John Yang and
                  Carlos E. Jimenez and
                  Alexander Wettig and
                  Kilian Lieret and
                  Shunyu Yao and
                  Karthik Narasimhan and
                  Ofir Press},
  editor       = {Amir Globersons and
                  Lester Mackey and
                  Danielle Belgrave and
                  Angela Fan and
                  Ulrich Paquet and
                  Jakub M. Tomczak and
                  Cheng Zhang},
  title        = {SWE-agent: Agent-Computer Interfaces Enable Automated Software Engineering},
  booktitle    = {Advances in Neural Information Processing Systems 38: Annual Conference
                  on Neural Information Processing Systems 2024, NeurIPS 2024, Vancouver,
                  BC, Canada, December 10 - 15, 2024},
  year         = {2024},
  url          = {http://papers.nips.cc/paper\_files/paper/2024/hash/5a7c947568c1b1328ccc5230172e1e7c-Abstract-Conference.html},
  bibsource    = {dblp computer science bibliography, https://dblp.org}
}

@article{crmweaver25lai,
  author       = {Yilong Lai and
                  Yipin Yang and
                  Jialong Wu and
                  Fengran Mo and
                  Zhenglin Wang and
                  Ting Liang and
                  Jianguo Lin and
                  Keping Yang},
  title        = {CRMWeaver: Building Powerful Business Agent via Agentic {RL} and Shared
                  Memories},
  journal      = {CoRR},
  volume       = {abs/2510.25333},
  year         = {2025},
  url          = {https://doi.org/10.48550/arXiv.2510.25333},
  doi          = {10.48550/ARXIV.2510.25333},
  eprinttype   = {arXiv},
  eprint       = {2510.25333},
  bibsource    = {dblp computer science bibliography, https://dblp.org}
}

@article{fama25yan,
  author       = {Yineng Yan and
                  Xidong Wang and
                  Jin Seng Cheng and
                  Ran Hu and
                  Wentao Guan and
                  Nahid Farahmand and
                  Hengte Lin and
                  Yue Li},
  title        = {FaMA: LLM-Empowered Agentic Assistant for Consumer-to-Consumer Marketplace},
  journal      = {CoRR},
  volume       = {abs/2509.03890},
  year         = {2025},
  url          = {https://doi.org/10.48550/arXiv.2509.03890},
  doi          = {10.48550/ARXIV.2509.03890},
  eprinttype   = {arXiv},
  eprint       = {2509.03890},
  bibsource    = {dblp computer science bibliography, https://dblp.org}
}

@article{autocodebench25chou,
  author       = {Jason Chou and
                  Ao Liu and
                  Yuchi Deng and
                  Zhiying Zeng and
                  Tao Zhang and
                  Haotian Zhu and
                  Jianwei Cai and
                  Yue Mao and
                  Chenchen Zhang and
                  Lingyun Tan and
                  Ziyan Xu and
                  Bohui Zhai and
                  Hengyi Liu and
                  Speed Zhu and
                  Wiggin Zhou and
                  Fengzong Lian},
  title        = {AutoCodeBench: Large Language Models are Automatic Code Benchmark
                  Generators},
  journal      = {CoRR},
  volume       = {abs/2508.09101},
  year         = {2025},
  url          = {https://doi.org/10.48550/arXiv.2508.09101},
  doi          = {10.48550/ARXIV.2508.09101},
  eprinttype   = {arXiv},
  eprint       = {2508.09101},
  bibsource    = {dblp computer science bibliography, https://dblp.org}
}

@article{deepscholarbench25patel,
  author       = {Liana Patel and
                  Negar Arabzadeh and
                  Harshit Gupta and
                  Ankita Sundar and
                  Ion Stoica and
                  Matei Zaharia and
                  Carlos Guestrin},
  title        = {DeepScholar-Bench: {A} Live Benchmark and Automated Evaluation for
                  Generative Research Synthesis},
  journal      = {CoRR},
  volume       = {abs/2508.20033},
  year         = {2025},
  url          = {https://doi.org/10.48550/arXiv.2508.20033},
  doi          = {10.48550/ARXIV.2508.20033},
  eprinttype   = {arXiv},
  eprint       = {2508.20033},
  bibsource    = {dblp computer science bibliography, https://dblp.org}
}

@inproceedings{evaluatingnovelty24merrill,
  author       = {William Merrill and
                  Noah A. Smith and
                  Yanai Elazar},
  editor       = {Yaser Al{-}Onaizan and
                  Mohit Bansal and
                  Yun{-}Nung Chen},
  title        = {Evaluating n-Gram Novelty of Language Models Using Rusty-DAWG},
  booktitle    = {Proceedings of the 2024 Conference on Empirical Methods in Natural
                  Language Processing, {EMNLP} 2024, Miami, FL, USA, November 12-16,
                  2024},
  pages        = {14459--14473},
  publisher    = {Association for Computational Linguistics},
  year         = {2024},
  url          = {https://doi.org/10.18653/v1/2024.emnlp-main.800},
  doi          = {10.18653/V1/2024.EMNLP-MAIN.800},
  bibsource    = {dblp computer science bibliography, https://dblp.org}
}

@article{standardizingdiversity24shaib,
  author       = {Chantal Shaib and
                  Joe Barrow and
                  Jiuding Sun and
                  Alexa F. Siu and
                  Byron C. Wallace and
                  Ani Nenkova},
  title        = {Standardizing the Measurement of Text Diversity: {A} Tool and a Comparative
                  Analysis of Scores},
  journal      = {CoRR},
  volume       = {abs/2403.00553},
  year         = {2024},
  url          = {https://doi.org/10.48550/arXiv.2403.00553},
  doi          = {10.48550/ARXIV.2403.00553},
  eprinttype   = {arXiv},
  eprint       = {2403.00553},
  bibsource    = {dblp computer science bibliography, https://dblp.org}
}

@inproceedings{evaluatingdiversity21tevet,
  author       = {Guy Tevet and
                  Jonathan Berant},
  editor       = {Paola Merlo and
                  J{\"{o}}rg Tiedemann and
                  Reut Tsarfaty},
  title        = {Evaluating the Evaluation of Diversity in Natural Language Generation},
  booktitle    = {Proceedings of the 16th Conference of the European Chapter of the
                  Association for Computational Linguistics: Main Volume, {EACL} 2021,
                  Online, April 19 - 23, 2021},
  pages        = {326--346},
  publisher    = {Association for Computational Linguistics},
  year         = {2021},
  url          = {https://doi.org/10.18653/v1/2021.eacl-main.25},
  doi          = {10.18653/V1/2021.EACL-MAIN.25},
  bibsource    = {dblp computer science bibliography, https://dblp.org}
}
\newpage
\appendix
\raggedbottom
\section{Additional Results}
\label{app:additional-results}

\paragraph{Coverage Desiderata.}
\label{app:sequence-metrics-full}
\begin{table}[H]
\centering
\caption{Sequence-level diagnostics. \emph{Gold} columns score the action lists in each task's evaluation criteria. \emph{GoldSim} columns score one representative agent sequence per task, selected as the shortest run with $\texttt{db\_reward}\!=\!1$ across all in-scope simulations; tasks with no successful run fall back to the gold action list. Selection rule and aggregation are detailed in Section~\ref{app:sequence-metrics-full}.}
\label{tab:sequence-metrics-goldsim}
\scriptsize
\setlength{\tabcolsep}{3pt}
\renewcommand{\arraystretch}{0.85}
\vspace{1mm}
\begin{tabular}{l cccc @{\hspace{0.6em}} cccc @{\hspace{0.6em}} cccc}
\toprule
 & \multicolumn{4}{c}{Airline} & \multicolumn{4}{c}{Retail} & \multicolumn{4}{c}{Telecom} \\
\cmidrule(lr){2-5}\cmidrule(lr){6-9}\cmidrule(lr){10-13}
 & \multicolumn{2}{c}{Gold} & \multicolumn{2}{c}{GoldSim} & \multicolumn{2}{c}{Gold} & \multicolumn{2}{c}{GoldSim} & \multicolumn{2}{c}{Gold} & \multicolumn{2}{c}{GoldSim} \\
\cmidrule(lr){2-3}\cmidrule(lr){4-5}\cmidrule(lr){6-7}\cmidrule(lr){8-9}\cmidrule(lr){10-11}\cmidrule(lr){12-13}
Metric & TBV & \textbf{Ours} & TBV & \textbf{Ours} & TBV & \textbf{Ours} & TBV & \textbf{Ours} & TBV & \textbf{Ours} & TBV & \textbf{Ours} \\
\midrule
\rowcolor[gray]{0.95}
Avg.\ length & 2.84 & 11.36 & 4.68 & 10.64 & 4.82 & 10.66 & 6.58 & 9.19 & 4.53 & 6.50 & 6.60 & 7.64 \\
W/R ratio & 0.53 & 0.31 & 0.27 & 0.37 & 0.47 & 0.32 & 0.30 & 0.40 & {--} & {--} & 0.35 & 0.68 \\
\rowcolor[gray]{0.95}
WED intra & 3.76 & 8.42 & 5.20 & 8.34 & 4.89 & 7.07 & 4.18 & 6.70 & 4.32 & 5.50 & 6.63 & 7.05 \\
\rowcolor[gray]{0.95}
Entropy 1-gram & 2.60 & 3.50 & 2.80 & 3.51 & 3.23 & 3.38 & 3.01 & 3.38 & 3.86 & 4.34 & 4.61 & 4.80 \\
Entropy 2-gram & 3.42 & 6.68 & 4.12 & 6.54 & 4.64 & 6.22 & 4.27 & 6.02 & 5.17 & 7.67 & 6.44 & 8.10 \\
\rowcolor[gray]{0.95}
Entropy 3-gram & 3.69 & 8.22 & 4.65 & 8.18 & 5.29 & 8.24 & 5.14 & 7.94 & 5.90 & 8.33 & 7.29 & 8.78 \\
Entropy 4-gram & 3.63 & 8.58 & 4.73 & 8.50 & 5.87 & 9.27 & 5.90 & 8.84 & 6.05 & 8.20 & 7.60 & 8.73 \\
\rowcolor[gray]{0.95}
Entropy 1-gram (norm.) & 0.68 & 0.92 & 0.74 & 0.92 & 0.83 & 0.86 & 0.77 & 0.86 & 0.71 & 0.80 & 0.85 & 0.88 \\
Entropy 2-gram (norm.) & 0.45 & 0.88 & 0.54 & 0.86 & 0.59 & 0.80 & 0.55 & 0.77 & 0.48 & 0.71 & 0.59 & 0.75 \\
\rowcolor[gray]{0.95}
Entropy 3-gram (norm.) & 0.32 & 0.72 & 0.41 & 0.72 & 0.45 & 0.70 & 0.44 & 0.68 & 0.36 & 0.51 & 0.45 & 0.54 \\
Entropy 4-gram (norm.) & 0.24 & 0.56 & 0.31 & 0.56 & 0.38 & 0.59 & 0.38 & 0.57 & 0.28 & 0.38 & 0.35 & 0.40 \\
\rowcolor[gray]{0.95}
Entropy avg.\ (norm.) & 0.42 & 0.77 & 0.50 & 0.76 & 0.56 & 0.74 & 0.53 & 0.72 & 0.46 & 0.60 & 0.56 & 0.64 \\
\# unique seqs & 30 & 50 & 40 & 50 & 75 & 114 & 94 & 112 & 94 & 114 & 107 & 102 \\
\rowcolor[gray]{0.95}
\# unique 2-gram & 20 & 133 & 41 & 135 & 65 & 127 & 64 & 132 & 57 & 279 & 149 & 374 \\
\# unique 3-gram & 24 & 333 & 56 & 322 & 92 & 418 & 108 & 386 & 86 & 365 & 228 & 494 \\
\rowcolor[gray]{0.95}
\# unique 4-gram & 23 & 393 & 54 & 368 & 105 & 680 & 130 & 560 & 93 & 317 & 255 & 451 \\
\# unique 5-gram & 18 & 362 & 47 & 334 & 103 & 725 & 139 & 552 & 80 & 244 & 234 & 371 \\
\rowcolor[gray]{0.95}
\# unique 6-gram & 14 & 315 & 42 & 288 & 86 & 644 & 135 & 491 & 60 & 171 & 200 & 288 \\
TTR 2-gram & 0.20 & 0.26 & 0.22 & 0.28 & 0.15 & 0.12 & 0.10 & 0.14 & 0.14 & 0.44 & 0.23 & 0.49 \\
\rowcolor[gray]{0.95}
TTR 3-gram & 0.32 & 0.71 & 0.38 & 0.74 & 0.27 & 0.42 & 0.21 & 0.47 & 0.28 & 0.71 & 0.43 & 0.77 \\
TTR 4-gram & 0.42 & 0.94 & 0.46 & 0.96 & 0.39 & 0.78 & 0.31 & 0.79 & 0.41 & 0.78 & 0.61 & 0.84 \\
\rowcolor[gray]{0.95}
TTR 5-gram & 0.44 & 0.98 & 0.51 & 1.00 & 0.51 & 0.96 & 0.44 & 0.92 & 0.49 & 0.80 & 0.71 & 0.86 \\
TTR 6-gram & 0.47 & 0.99 & 0.55 & 1.00 & 0.61 & 0.99 & 0.60 & 0.98 & 0.56 & 0.81 & 0.78 & 0.88 \\
\rowcolor[gray]{0.95}
\# unique avg.\ (TTR) & 0.37 & 0.78 & 0.43 & 0.80 & 0.39 & 0.65 & 0.33 & 0.66 & 0.38 & 0.71 & 0.55 & 0.77 \\
\bottomrule
\end{tabular}
\end{table}

Table~\ref{tab:sequence-metrics-goldsim} extends the results presented in Figure~\ref{fig:coverage_fig} to the full set of sequence-level statistics: average length (mean number of tool calls per sequence); write/read balance (ratio of write tools to read and generic tools); within-cluster weighted edit distance (\emph{WED intra} - mean pairwise distance over all sequence pairs in the cluster, indicating intra-set diversity); Shannon entropy at \mbox{$n=1\ldots4$} (n-gram occurrences are tallied with a sliding window, normalised to a probability distribution, then $H_n=-\sum_g p(g)\log_2 p(g)$) together with their vocabulary-normalised mean ($H_n/(n\log_2|V|)$ averaged   over $n$, in $[0,1]$, allowing comparison across cells with different tool vocabularies); the count of unique $n$-grams at $n=2\ldots6$ (compositional richness); and the mean type-token ratio (unique $n$-grams divided by total $n$-grams positions, averaged over $n=2\ldots6$; higher means less repetition). The reported numbers compare against \emph{GoldSim}; how we extract it is explained in the following paragraph.

\paragraph{The GoldSim simulation set.}
The \emph{GoldSim} columns aggregate agent behaviour into one representative tool sequence per task. Concretely, for a given (domain, task-set) cell we proceed as follows:
\begin{enumerate}
    \item \textbf{Run pool.} For every task $i$, gather the union of all simulation runs, pooled across agent, user simulator, and trial $k$. Filter unsuccessful runs (reward$=0.0$).
    \item \textbf{Sequence extraction.} For each run we read the agent's tool calls in turn order and concatenate their identifiers into one ordered tool sequence ($\bar\sigma$).
    \item \textbf{Per-task pick.} If more then a single candidate exist, we pick the candidate of \emph{minimal length}. The intuition is that the shortest successful trajectory is the cleanest evidence of a working solution path.
    \item \textbf{Gold fallback.} If no candidate exists for task $i$ ---  we use the task's gold tool sequence ($\bar\sigma^\star$). This keeps the per-cell sequence count equal to the cell's task count and avoids selection bias against hard tasks.
\end{enumerate}

\paragraph{Difficulty Desiderata.}
\label{app:difficulty-desiderata}
\begin{figure}[H]
    \centering
    \includegraphics[width=0.7\linewidth]{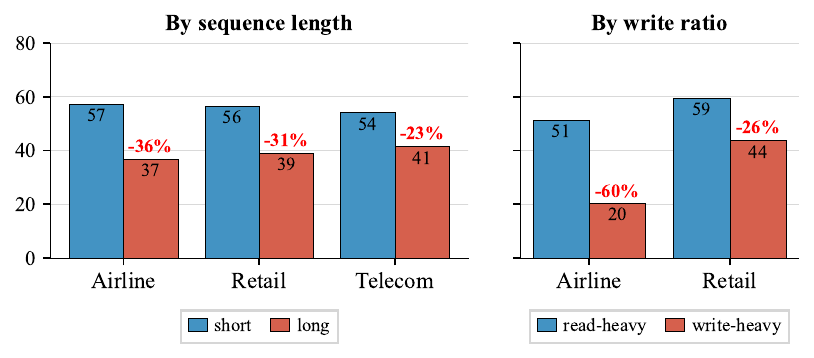}
    \caption{Mean accuracy on tasks bucketed by per-task golden-sequence length (left) and write-ratio (right), using $\mu \pm 0.5\sigma$ thresholds with the middle band dropped.}
    \label{fig:difficulty-desiderata}
\end{figure}

Figure~\ref{fig:difficulty-desiderata} validates the two structural axes we use to control task difficulty. Splitting tasks by per-task golden-sequence length produces a $13$--$20$\,pp accuracy gap between short and long buckets across all three domains. Splitting by per-task write-ratio produces a $16$--$31$\,pp gap between read-heavy and write-heavy buckets in airline and retail. Telecom is excluded from the write-ratio panel because its evaluation criteria encode only write actions, so per-task write-ratio collapses to $1.0$ for every task. These results indicate that compositional length and write-heaviness act as genuine difficulty knobs rather than descriptive labels, supporting their use as control axes when sampling harder tasks.

\begin{figure}[H]
\centering

\begin{minipage}[t]{0.49\linewidth}
    \centering
    \vspace*{0pt}%
    \includegraphics[width=\linewidth]{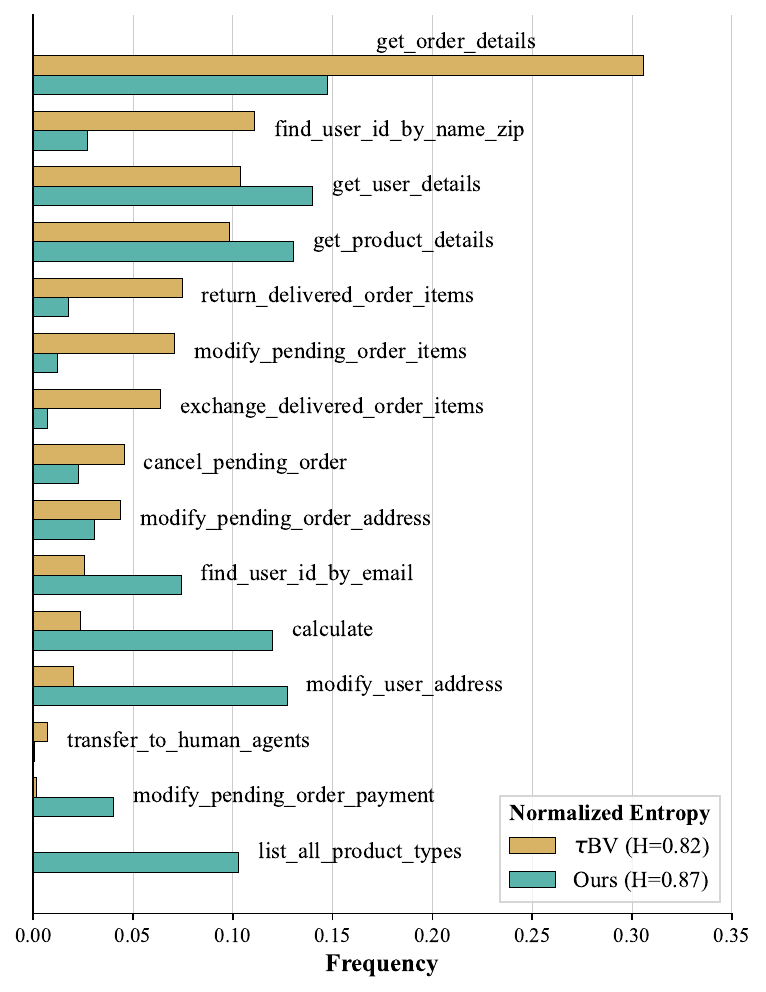}\\
    \textbf{(a) retail}
\end{minipage}%
\hfill
\begin{minipage}[t]{0.49\linewidth}
    \centering
    \vspace*{0pt}%
    \includegraphics[width=\linewidth]{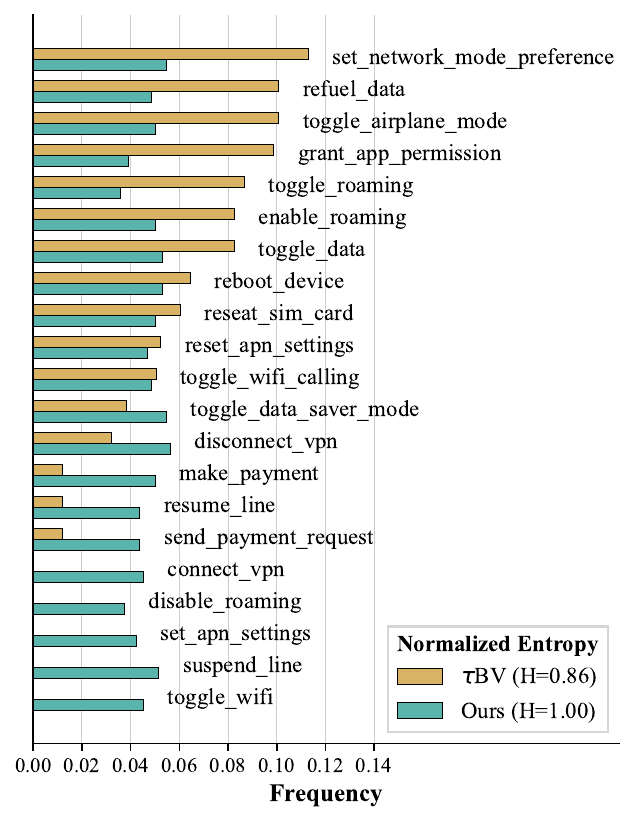}\\
    \textbf{(b) telecom (write actions only)}
\end{minipage}

\caption{Per-tool relative frequency on the retail and telecom domains, comparing \tbv\xspace against \ourbenchmark; the airline counterpart is in Figure~\ref{fig:coverage_fig}. Normalised entropy $H$ is reported in each panel's legend; tools are ordered within each panel by \tbv\xspace task frequency. Telecom is restricted to write actions (21 of 43 tools, both distributions re-normalised over this subset) since the \tbv\xspace telecom tasks does not include read tools in their gold tool sequences.}
\label{fig:diversity-combined}
\end{figure}

\paragraph{Tool-Usage Distributions Across Domains.}
\label{app:tool-use-dist}
Figure~\ref{fig:diversity-combined} reports complementary diversity views across the three \taubench domains. Each plot the per-tool relative frequency of the \tbv\xspace against \ourbenchmark; the latter spreads usage more evenly across the tool vocabulary, raising the normalised entropy ($H$ = Shannon entropy / $\log_2 |\text{tools}|$) reported in each legend.

\paragraph{Qualitative Comparison: Weighted vs.\ Regular Edit Distance.}
\label{app:wed-qualitative}
Figure~\ref{fig:cluster-distances} (Appendix~\ref{app:examples}) contrasts $k$-medoids clustering on airline trajectories under our weighted edit distance (WED) and the unweighted edit distance (ED). Under WED \textbf{(a)}, the three medoids correspond to clear, distinct user-service intents -- \emph{search/book/update}, \emph{demand a refund}, and \emph{book and update} -- and each medoid's nearest neighbors stay close to it semantically: they share the medoid's overall intent and only differ in surface-level details. Under ED \textbf{(b)}, the medoids themselves remain interpretable, but the nearest neighbors drift away from the medoid's intent, mixing in unrelated tool subsequences.  Together, these examples illustrate why we use WED rather than plain edit distance for the clustering step in Section~\ref{sub:clustering}: the type-aware substitution costs translate into clusters whose members are semantically coherent, not just superficially similar.

\section{Supplemental Direct-LLM Sequence Comparison}
\label{app:direct-llm}

We compare $100$ Airline tool-name sequences directly generated by GPT-6 Luna with $100$ samples from TASTE's trained n-gram model. The direct generator receives the Airline policy, tool definitions, and original gold sequences. Both methods use the same sequence lengths distribution. Table~\ref{tab:direct-llm} shows the same coverage metrics reported in Section~\ref{sec:results}: mean pairwise WED, TTR averaged over $n=2,\ldots,6$, and normalized tool-frequency entropy. In this run, TASTE produces more distinct sequences and higher WED and TTR, while direct prompting has higher tool-frequency entropy. This comparison concerns unfiltered sequences; plausibility and complete task generation were not evaluated.

\begin{table}[H]
\centering
\caption{Airline sequence comparison using the paper's existing coverage metrics. Each method produces $100$ sequences with identical lengths. Entropy is normalized by $\log_2|\mathcal{T}|$.}
\label{tab:direct-llm}
\small
\begin{tabular}{@{}lrr@{}}
\toprule
Metric & Direct GPT & TASTE \\
\midrule
Unique sequences & 56 & 68 \\
Mean pairwise WED & 3.34 & 3.51 \\
TTR (mean, $n=2,\ldots,6$) & 0.49 & 0.84 \\
Tool-frequency entropy (norm.) & 0.904 & 0.865 \\
\bottomrule
\end{tabular}
\end{table}

 \section{Implementation Details}
\label{app:method_implementation}

This appendix documents the end-to-end pipeline of TASTE (Section~\ref{sec:method}), as executed to produce \ourbenchmark. The description is domain-agnostic: the only domain-specific resources are environment definition (policy document, tool specification, DB schema) and the set of original tasks.

All three stages are executed as separate processes that communicate via artifacts; each stage's output is the following stage's input. Unless stated otherwise, every LLM call uses Gemini-$3$-flash with default LiteLLM\footnote{\url{https://www.litellm.ai/}} settings; ablations over the model choice are reported in the Results section (Section~\ref{sec:results}).

\subsection{Stage 1: Tool Sequence Sampling with an Adaptive Contrastive $n$-gram Model}
\label{app:n-gram}

Algorithm~\ref{alg:ngram-training} summarizes the procedure described below.

\paragraph{Model.}
The sampler instantiates the Adaptive Contrastive $n$-gram model of Section~\ref{sub:n_gram} with $n{=}3$ (trigram), Dirichlet smoothing $\lambda_0 = 0.1$, and negative-evidence weight $\lambda_{\text{neg}} = 1$. Positive and negative count tables, $C^+$ and $C^-$, are initially empty. The available tokens set is the environment's tool identifiers; two special tokens, \textsc{bos} and \textsc{eos}, are reserved for padding and termination and are never emitted by the sampler.

\paragraph{Temperature schedule.}
The temperature schedule is instantiated as an exponential decay
\[
  T(k) \;=\; 1 + (T_0 - 1)\,\exp\!\bigl(-k / \tau_{\text{decay}}\bigr),
\]
with $T_0 = 3.0$ and $\tau_{\text{decay}} = 1500$. The iteration counter $k$ advances once per generation attempt.

\paragraph{Length sampling.}
Candidate sequence lengths are drawn from a skew-normal distribution with location~$7$, scale~$5$, and skewness~$2.0$; samples are hard-clipped to the integer range $[1, 15]$ and rounded to the nearest integer.
 The $15$-call cap is an experimental configuration, not an architectural limit; longer gold sequences remain unevaluated, and evolution preserves the gold-sequence horizon.

\paragraph{Generation.}
A candidate action sequence is produced by prepending two \textsc{bos} tokens to the empty sequence, then sampling tokens autoregressively from $P(t_i \mid \text{ctx}_i)$ until the length $L$ drawn above is reached;

\paragraph{Plausibility validator.}
Every candidate is evaluated by the plausibility validator: an LLM prompted with the domain-specific validation template (Box~\ref{box:validate-action-sequence}). The prompt is templated with the candidate sequence and the domain environment. The validator determines whether the candidate sequence is valid and produces problematic indices if it is not.

\paragraph{Ingestion.}
Sequences judged plausible are ingested into $C^+$: every $n$-gram window extracted from the \textsc{bos}-padded sequence is incremented. Sequences judged implausible are ingested into $C^-$: only windows whose final token index appears in problematic indices are incremented.  Uniqueness is enforced by tuple-equality deduplication. Duplicate sequences are skipped without contributing to either table; the iteration counter~$k$ still advances.

\paragraph{Seed initialization.}
Before the main loop, the gold tool sequence of the domain's original tasks are each evaluated by the plausibility validator and ingested according to its verdict.

\paragraph{Training loop.}
The loop runs for a total of $3{,}000$ generation attempts. The output of Stage~1 is the trained sampler with its final count tables.

\subsection{Stage 2: Clustering and Selection}
\label{app:clustering}

Algorithm~\ref{alg:clustering} summarizes the procedure described below.

\paragraph{Pool generation.}
A pool of $2{,}000$ unique tool sequences is drawn from the trained sampler. Sampling uses a fixed temperature $T_{\text{pool}} = 1.5$ and the same skew-normal length distribution as in Stage~1.

\paragraph{Weighted edit distance.}
Pairwise distances between sequences use a standard Levenshtein recurrence with insertion and deletion cost $1$ and a hand-set tiered substitution cost:
\[
  d_{\text{sub}}(a, b) \;=\;
  \begin{cases}
    0 & \text{if } a = b, \\
    0.33 & \text{if } \text{type}(a) = \text{type}(b) \text{ and } \text{group}(a) = \text{group}(b), \\
    0.66 & \text{if } \text{type}(a) = \text{type}(b) \text{ and } \text{group}(a) \neq \text{group}(b), \\
    1.00 & \text{if } \text{type}(a) \neq \text{type}(b).
  \end{cases}
\]
The \emph{type} of a tool is one of \textsc{read}, \textsc{write}, \textsc{generic}, as declared in the tool specification; \textsc{generic} tools are treated as \textsc{read} for the type comparison only. The \emph{group} of a tool is the prefix of the tool name up to the first underscore (e.g., the group of \texttt{search\_direct\_flight} is \texttt{search}).

\begin{figure}[H]
\centering

\begin{minipage}[t]{\linewidth}
    \centering
    \includegraphics[width=0.98\linewidth]{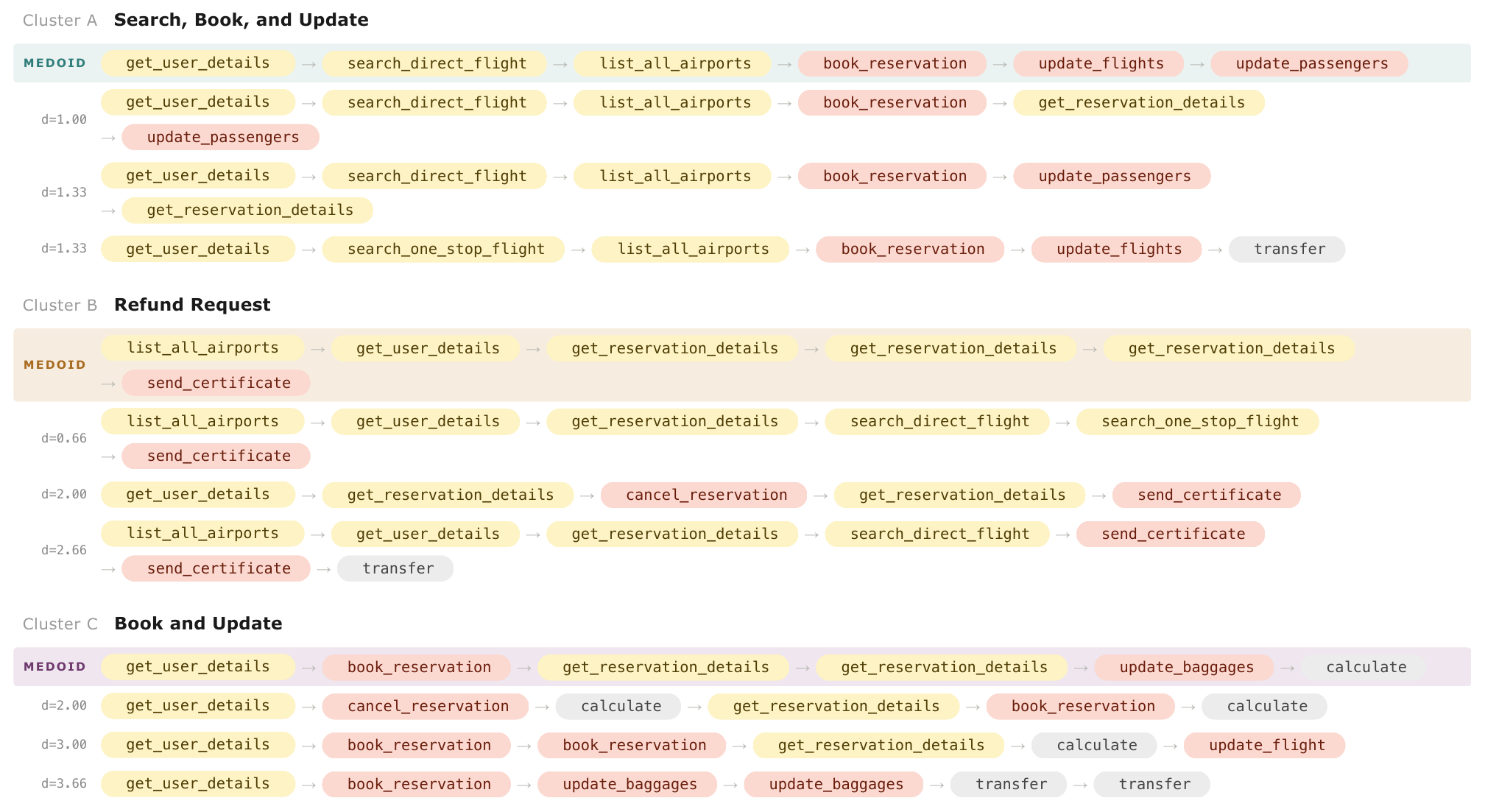}\\
    \textbf{(a)}
\end{minipage}\\[0.4em]
\begin{minipage}[t]{\linewidth}
    \centering
    \includegraphics[width=0.98\linewidth]{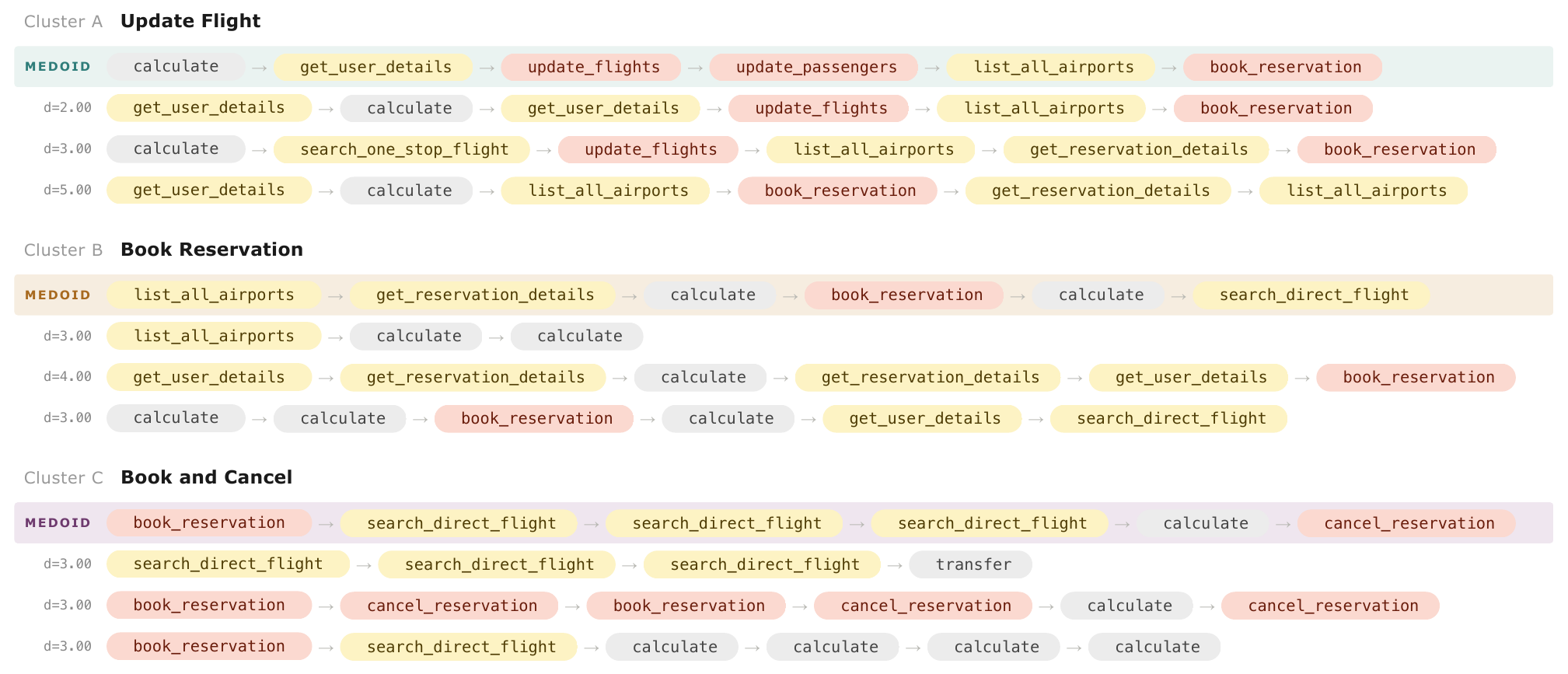}\\
    \textbf{(b)}
\end{minipage}

\caption{$k$-medoids clusters on the airline domain under weighted edit distance \textbf{(a)} vs.\ unweighted edit distance \textbf{(b)}. Each cluster shows its medoid (top row) and three nearest neighbors; chips are colored by tool type (\textsc{read}, \textsc{write}, \textsc{generic}).}
\label{fig:cluster-distances}
\end{figure}

Figure~\ref{fig:cluster-distances} (Appendix~\ref{app:wed-qualitative}) gives a qualitative illustration of how this metric behaves on real airline-domain sequences. For three clusters drawn from the validated $k=50$ medoid set with visibly distinct user-service intents (book a new reservation; modify reservation flights; cancel a reservation with a refund certificate), Figure~\ref{fig:cluster-distances}(a) shows that within-cluster distances stay between $1.0$ and $3.98$, while the inter-medoid matrix in Figure~\ref{fig:cluster-distances}(b) reports a smallest between-medoid distance of $5.31$ -- so the cluster boundaries align with semantic intent, and cosmetic variants (read reordering, optional same-group write substitutions) are absorbed into the same cluster.

\paragraph{$K$-medoids.}
The number of clusters $K$ is set to the number of seed tasks in the domain ($50$ for airline; $114$ for retail and telecom). Medoids are initialized via the $k$-medoids\texttt{++} scheme: the first medoid is drawn uniformly at random from the pool, and each subsequent medoid is drawn with probability proportional to the squared distance from the pool point to its nearest existing medoid. A PAM-style update loop then alternates an assignment step (each point is assigned to its nearest medoid) with an update step (each medoid is replaced by the cluster member that minimizes total intra-cluster distance). The loop terminates when the medoid set stabilizes between consecutive iterations, with a hard cap of $100$ iterations.

\paragraph{Medoid validation and replacement.}
Each medoid is validated by the same plausibility validator used in Stage~1. When a medoid is judged implausible, a replacement is searched among the other members of its cluster in order of increasing distance to the original medoid. If no member of the cluster is judged plausible, the cluster is marked \emph{unusable} for this round.

\paragraph{Re-clustering of unusable clusters.}
At the start of each re-clustering round, every medoid that has been accepted in a previous round is frozen as an immovable center. A fresh pool of sequences is drawn from the sampler under the same settings as the initial pool. Constrained $k$-medoids is then run with the frozen centers fixed and $k'$ free centers (one per unusable cluster) selected from the new pool; the constraint is algorithmic rather than penalty-based: frozen indices never participate in the medoid-update step, but they remain in the nearest-medoid assignment. Newly selected free medoids are validated as above; unusable clusters remaining after three rounds are dropped, and the benchmark size is reduced accordingly.

\paragraph{Output.}
The stage emits, for each plausible cluster, the medoid sequence.

\subsection{Stage 3: Task Generation and Evolution}
\label{app:task-generation}

Each plausible medoid $\bar{\sigma}$ from Stage~2 is instantiated as a base task $x=(s_0, u, \sigma^\star)$ following Section~\ref{sub:task_generation}: two LLM calls produce the task, which is then screened by a multi-step validity pipeline with a feedback-driven repair loop. The validated base task is then evolved into an adversarial variant by a three-step pipeline with its own validity check. Algorithm~\ref{alg:task-generation} summarizes the procedure described below.

\paragraph{Base task generation: scenario construction.}
The first LLM call (Box~\ref{box:create-user-task}) constructs a coherent real-world scenario that motivates the tool sequence $\bar{\sigma}$, inventing concrete entities and producing a verbose, step-by-step user instruction $u$.

\paragraph{Base task generation: database initialization.}
The second LLM call (Box~\ref{box:generate-db-init}) produces the database entities required to execute $\bar{\sigma}$. These entities are merged into the domain's base environment state to form the task's initial state $s_0$.

\paragraph{Validity checks.}
Following Section~\ref{sub:task_generation}, the generated task is evaluated by a pipeline combining rule-based and simulation-based checks; the pipeline short-circuits on the first failure.
\begin{enumerate}
  \item \textbf{(Rule)} Arguments fully specified.
  \item \textbf{(Rule)} Referenced entities exist in $s_0$.
  \item \textbf{(Rule)} $s_0$ conforms to the domain schema.
  \item \textbf{(Rule)} Gold tool-call sequence succeeds.
  \item \textbf{(LLM)} Policy coherence review. An LLM reviews the task instructions against the policy (Box~\ref{box:task-coherence-review}).
  \item \textbf{(Simulation)} Verifier agent. Described below.
\end{enumerate}

\paragraph{Verifier agent.}
\label{app:verifier-hints}
The verifier agent of Section~\ref{sub:task_generation} is built on \taubench's \texttt{LLMGTAgent} workflow, which primes an LLM with the full list of gold tool calls (``action hints'') before running the task in \taubench's standard simulator. Running the simulation with fully-specified hints might be too easy: the agent can transcribe the tool calls into the conversation without engaging with the scenario, yields a degenerate solvability check. Running it with no hints is too hard: a competent agent may fail a perfectly valid task because the task itself is difficult. We therefore deviate from the base implementation in two places, designed to leave enough scaffolding for the task to be solvable while forcing the agent to ground each decision in the conversation.

We modify the hint list   by masking argument values in selected gold calls and optionally shuffling their order. The masking fraction and hint order are configurable.  A task passes if the hinted-simulation yield a reward of $1.0$.

\paragraph{Feedback-driven repair and cluster replacement.}
When any validity check fails, the pipeline invokes a repair mechanism that feeds the failing step's error back to an LLM, which attempts to patch the offending part of the task or database. Tasks that cannot be repaired are regenerated from scratch; if repeated attempts fail, the medoid is replaced by another candidate from its cluster (Section~\ref{sub:clustering}).

\paragraph{Task evolution: strategy analysis.}
Each validated base task is then evolved into a harder variant. The first evolution LLM call (Box~\ref{box:adv-strategy}) examines each \textsc{write} action and selects global techniques from the catalog described below that shape the conversation as a whole, defines conditional branches describing how the user reacts to different agent responses, and lists decoy entities to be constructed in the environment-perturbation step.

\paragraph{Adversarial pattern catalog.}
\label{app:adversarial-patterns}
We define a catalog of adversarial patterns, each targeting a distinct failure mode of tool-using agents. We divide the catalog into three main categories, each holding a few applicable patterns. The strategy phase selects a pattern per write action, based on the specific action and database context.
\begin{enumerate}
    \item \textbf{DB-grounded misdirection.} The user leads the agent towards plausible but wrong argument values. A careful agent catches these by looking things up in the DB. e.g., Decoy entity - where a similar-but-wrong entity exists in the DB that plausibly matches the user's vague description, with subtle disqualifier.
    \item \textbf{Policy Enforcement.} The user pushes the agent toward actions that policy forbids, or beyond what the task requires. e.g., Eligibility Pressure - where a user demands an action that policy forbids, such as ineligible refund.
    \item \textbf{Conversational adversity.} The user controls the flow of information to stress-test the agent's patience and robustness. e.g., Information withholding.
\end{enumerate}

\paragraph{Task evolution: environment perturbation.}
A second LLM call materializes the designed techniques into concrete database records, which are merged into $s_0$ additively.

\paragraph{Task evolution: scenario rewriting.}
A third LLM call (Box~\ref{box:adv-scenario}) rewrites the user instruction $u$ to deploy the adversarial techniques from the strategy step and exploit the decoys from the environment-perturbation step, while keeping $\bar{\sigma}$ and its gold arguments unchanged.

\paragraph{Evolved-task validation and graduated fallback.}
The evolved task passes through the validation pipeline mentioned above; Rule-based steps are skipped, as $\bar{\sigma}$, its arguments, and the environment outside the decoys are preserved by construction. Failures at any evolution step trigger a feedback-driven retry with the failing step's error surfaced to the LLM. If full adversarial evolution still fails, the graduated fallback described below applies: a \emph{lite} rewrite is attempted without strategy analysis or decoys, and if it also fails the original base task is retained unchanged.

\paragraph{Graduated fallback.}
\label{app:graduated-fallback}
When full adversarial evolution fails validation after multiple retry attempts, the pipeline applies a \emph{lite} fallback: a simplified evolution that adds only mild adversarial elements --- without strategy analysis. If the lite version also fails, the original base task is retained unchanged. This graduated approach maximizes the fraction of tasks that receive some level of difficulty enhancement while ensuring that the final task set is always complete.

\paragraph{Output.}
The stage emits a set of tasks conforming to the $\tau^2$-bench task schema, one per cluster, each in its evolved form.

\subsection{Consolidated Hyperparameter Table}
\label{app:impl-hparams}
Table~\ref{tab:hparams} reports the hyperparameters used in the reported runs; values are constant across domains unless noted. Table~\ref{tab:model-identifiers} lists the exact API identifiers behind every model name used in the paper.
 AppWorld settings are given in Appendix~\ref{app:appworld}.
\begin{table}[H]
\centering
\caption{Hyperparameters used in the reported runs.}
\label{tab:hparams}
\small
\setlength{\tabcolsep}{4pt}
\begin{minipage}[t]{0.49\textwidth}
\centering
\begin{tabular}{@{}ll@{}}
\toprule
Hyperparameter & Value \\
\midrule
\multicolumn{2}{@{}l}{\textit{Stage 1: Sampler}} \\
$n$-gram order $n$                        & $3$ \\
Dirichlet smoothing $\lambda_0$           & $0.1$ \\
Neg.-evidence weight $\lambda_{\text{neg}}$ & $1.0$ \\
Initial temperature $T_0$                 & $3.0$ \\
Temp.\ decay $\tau_{\text{decay}}$        & $1500$ \\
Length dist.\ (loc, scale, $\alpha$)      & skew-normal $(7, 5, 2.0)$ \\
Length clip                               & $[1, 15]$ \\
Training steps                            & $3000$ \\
\bottomrule
\end{tabular}
\end{minipage}%
\hfill
\begin{minipage}[t]{0.49\textwidth}
\centering
\begin{tabular}{@{}ll@{}}
\toprule
Hyperparameter & Value \\
\midrule
\multicolumn{2}{@{}l}{\textit{Stage 2: Clustering and Selection}} \\
Pool size $N$                             & $2000$ \\
Pool temperature $T_{\text{pool}}$        & $1.5$ \\
$K$-medoids max iterations                & $100$ \\
\midrule
\multicolumn{2}{@{}l}{\textit{Stage 3: Task Gen.\ \& Evolution}} \\
Verifier-agent redaction $p$                & $0.3$ \\
Verifier-agent simulation attempts          & $2$ \\
Lite fallback retries                     & $3$ \\
\bottomrule
\end{tabular}
\end{minipage}
\end{table}
\begin{table}[H]
\centering
\caption{Mapping between model names used throughout the paper and the exact API identifiers used for generation and evaluation. PPT (price per task) reports the average cost in USD of running one task with the given model.}
\label{tab:model-identifiers}
\small
\setlength{\tabcolsep}{6pt}
\vspace{2mm}
\begin{tabular}{@{}lllr@{}}
\toprule
Name in paper & Provider & API identifier & PPT (\$) \\
\midrule
Gemini-3-Flash    & Google     & \texttt{gemini-3-flash-001}        & 0.077 \\
Gemini-2.5-Flash  & Google     & \texttt{gemini-2.5-flash-001}      & 0.036 \\
GPT-5.2           & OpenAI     & \texttt{gpt-5.2-2025-12-11}        & 0.091 \\
Claude-Sonnet-4.6 & Anthropic  & \texttt{claude-4-6-sonnet-20260217} & 0.490 \\
DeepSeek-3.1      & DeepSeek   & \texttt{deepseek/deepseek-chat-v3.1} & 0.034 \\
Qwen-32B          & Alibaba    & \texttt{qwen/qwen-3-32b-instruct}  & 0.020 \\
\bottomrule
\end{tabular}
\end{table}

\subsection{Pseudo-codes}
\label{app:pseudo-codes}
\begin{algorithm}[h]
\caption{Adaptive contrastive $n$-gram training (Section~\ref{sub:n_gram}).}
\label{alg:ngram-training}
\small
\begin{algorithmic}[1]

\Require seed sequences $\mathcal{\sigma}_{\text{seed}}$, validator $V$, length dist.\ $L$, temperature $T(\cdot)$, iterations $K_{\text{iter}}$
\State $C^+, C^- \op{=} \emptyset$

\For{$\bar\sigma \in \mathcal{X}_{\text{seed}}$} \Comment{Initialize with seed tasks}
    \State $v, P \op{=} V(\bar\sigma)$
    \If{$v \op{==} \text{plausible}$}
        \State $C^+ \op{+=} \fn{ngrams}(\bar\sigma)$
    \Else
        \State $C^- \op{+=} \fn{ngrams}(\bar\sigma, P)$
    \EndIf
\EndFor
\For{$k \op{=} 1 \dots K_{\text{iter}}$} \Comment{Training loop}
    \State $\ell \op{\sim} L$
    \State $\bar\sigma \op{=} \fn{sample}(C^+, C^-, \ell, T(k))$
    \If{$\bar\sigma \op{\in} C^+\cup C^-$} \State \kw{continue} \EndIf
    \State $v, P \op{=} V(\bar\sigma)$
    \If{$v \op{==} \text{plausible}$}
        \State $C^+ \op{+=} \fn{ngrams}(\bar\sigma)$
    \Else
        \State $C^- \op{+=} \fn{ngrams}(\bar\sigma, P)$
    \EndIf
\EndFor
\State \kw{return} $C^+, C^-$

\end{algorithmic}
\end{algorithm}

\begin{algorithm}[h]
\caption{Clustering and medoid validation (Section~\ref{sub:clustering}).}
\label{alg:clustering}
\small
\begin{algorithmic}[1]

\Require trained $C^{\pm}$, validator $V$, target $K$, pool $N$, length dist.\ $L$, temp. $T_{\text{pool}}$, distance $d_w$, rounds $R_{\max}$
\State $\mathcal{M}_{\text{valid}} \op{=} \emptyset$
\For{$r \op{=} 1 \dots R_{\max}$}
    \State $\mathcal{P} \op{=} \emptyset$
    \While{$|\mathcal{P}| \op{<} N$} \Comment{Pool generation}
        \State $\ell \op{\sim} L$
        \State $\bar\sigma \op{=} \fn{sample}(C^+, C^-, \ell, T_{\text{pool}})$
        \If{$\bar\sigma \op{\notin} \mathcal{P}$}
            \State $\mathcal{P}.\fn{add}(\bar\sigma)$
        \EndIf
    \EndWhile
    \State $\{(\mathcal{C}_j, m_j)\}_{j=1}^{K} \op{=} \fn{k\_medoids}(\mathcal{P}, K, d_w, \fn{frozen}\op{=}\mathcal{M}_{\text{valid}})$ \Comment{Clustering}
    \For{$j \op{=} 1 \dots K$}
        \If{$m_j \op{\in} \mathcal{M}_{\text{valid}}$} \State \kw{continue} \EndIf
        \State $\mathcal{C}_j \op{=} \fn{sorted}(\mathcal{C}_j, \fn{key}\op{=} d_w(\cdot, m_j))$
        \For{$\bar\sigma \op{\in} \mathcal{C}_j$} \Comment{Medoids validation}
            \If{$V(\bar\sigma) \op{==} \text{plausible}$}
                \State $\mathcal{M}_{\text{valid}}.\fn{add}(\bar\sigma)$
                \State \kw{break}
            \EndIf
        \EndFor
    \EndFor
    \If{$|\mathcal{M}_{\text{valid}}| \op{==} K$} \State \kw{break} \EndIf
\EndFor
\State \kw{return} $\mathcal{M}_{\text{valid}}$

\end{algorithmic}
\end{algorithm}

\begin{algorithm}[h]
\caption{Base task generation and evolution (Section~\ref{sub:task_generation}).}
\label{alg:task-generation}
\small
\begin{algorithmic}[1]

\Require validated medoids $\mathcal{M}_{\text{valid}}$, task validator $V$, rounds $R^b_{\max}$, rounds $R^e_{\max}$, rounds $R^l_{\max}$
\State $\mathcal{X} \op{=} \emptyset$
\For{$\bar\sigma \op{\in} \mathcal{M}_{\text{valid}}$}
    \State $\text{valid\_base} = \kw{False}$
    \For{$r \op{=} 1 \dots R^b_{\max}$} \Comment{Construct base task}
        \State $u, \sigma^\star \op{=} \fn{llm\_scenario}(\bar\sigma)$
        \State $s_0 \op{=} \fn{llm\_init\_db}(\bar\sigma, u)$
        \State $x_{\text{base}} \op{=} (s_0, u, \sigma^\star)$
        \If{$V(x_{\text{base}}) \op{==} \text{pass}$} \Comment{Base task is valid}
        \State $\text{valid\_base} \op{=} \kw{True} $ 
        \State \kw{break}
        \EndIf
    \EndFor
    \If{$\text{valid\_base} \op{==} \kw{False}$}
    \State $\fn{mark\_base\_generation\_failed}(\bar\sigma)$ \Comment{Mark medoid for later re-clustering}
    \State \kw{continue}
    \EndIf
    \Statex
    \State $\text{evolved} \op{=} \kw{False}, \text{lite} \op{=} \kw{False}$
    \For{$r \op{=} 1 \dots R^e_{\max}$} \Comment{Evolve task}
        \State $x_{\text{evolved}} = \fn{evolve\_with\_llm}(x_{\text{base}}, \kw{mode} \op{=} \text{full})$
        \If{$V(x_{\text{evolved}}) \op{==} \text{pass}$}
            \State $\text{evolved} \op{=} \kw{True}$
            \State \kw{break}
        \EndIf
    \EndFor
    \If{$\text{evolved}$}
        \State $\mathcal{X}.\fn{add}(x_{\text{evolved}})$
        \State \kw{continue}
    \EndIf
    \For{$r \op{=} 1 \dots R^l_{\max}$} \Comment{Evolve failed, try lite evolvement}
        \State $x_{\text{lite}} = \fn{evolve\_with\_llm}(x_{\text{base}}, \kw{mode}\op{=}\text{lite})$
        \If{$V(x_{\text{lite}}) \op{==} \text{pass}$}
            \State $\text{lite} = \kw{True}$
            \State \kw{break}
        \EndIf
    \EndFor
    \If{$\text{lite}$}
        \State $\mathcal{X}.\fn{add}(x_{\text{lite}})$
        \State \kw{continue}
    \EndIf
    \State $\mathcal{X}.\fn{add}(x_{\text{base}})$ \Comment{lite failed, keep base task}
\EndFor
\State \kw{return} $\mathcal{X}$

\end{algorithmic}
\end{algorithm}

\subsection{AppWorld Adaptation and Evaluation}
\label{app:appworld}

We apply TASTE's sampling and clustering stages to the spec of API-call sites extracted from $147$ AppWorld train/dev gold programs. We used GPT-5.4 to realize selected sequences as $50$ base tasks, with executable gold programs and task-specific answer. Gemini-3-Flash and Claude-Sonnet-4.6 served as hardening targets.

Table~\ref{tab:appworld-results} reports task goal completion from one run per model and task set using AppWorld's official code agent, with a $50$-step limit. Base and Hardened use the same $50$ clusters; Original is a separate $24$-task \texttt{test\_normal} subset taken from official release.

\begin{table}[H]
\centering
\caption{\textbf{AppWorld extension:} Task goal completion (\%). Base and Hardened contain $50$ tasks each from the same clusters; Original contains $24$ separate tasks. $\Delta\%$ is relative to Base (Appendix~\ref{app:appworld}).}
\label{tab:appworld-results}
\small
\begin{tabular}{lrrrr}
\toprule
Agent & Original & Base & Hardened & $\Delta\%$ \\
\midrule
Claude-Sonnet-4.6 & 87.5 & 86.0 & 56.0 & $-34.9$ \\
Gemini-3-Flash   & 75.0 & 84.0 & 50.0 & $-40.5$ \\
DeepSeek-V4-Flash             & 79.2 & 84.0 & 60.0 & $-28.6$ \\
DeepSeek-3.1                  & 41.7 & 84.0 & 54.0 & $-35.7$ \\
\bottomrule
\end{tabular}
\par\smallskip
\end{table}

\section{Examples}
\label{app:examples}

\paragraph{Base and Evolved Tasks.}
\label{app:full-task-examples}
Figure~\ref{fig:full-task-examples} shows one end-to-end task example from each of the three \taubench domains, illustrating the output of the task generation and evolution pipeline of Section~\ref{sub:task_generation}. For each task we render the gold tool-call sequence, the \emph{base} variant, and the corresponding \emph{evolved} variant stacked on a single page. The task purpose is shown above the gold tool sequence and it is shared across both variants. The comparison between variants is shown across the \emph{Known Info} and the \emph{Task Instructions}.

\newcommand{\tasksamplefig}[1]{%
  \includegraphics[width=\linewidth,height=0.92\textheight,keepaspectratio,page=#1]{assets/figures/examples/base_vs_evolve_examples.pdf}%
}

\begin{figure}[p]
\centering
\tasksamplefig{1}
\caption{\textbf{(a)} airline.}
\label{fig:full-task-examples}
\end{figure}

\begin{figure}[p]\ContinuedFloat
\centering
\tasksamplefig{2}
\caption[]{\textbf{(b)} retail.}
\end{figure}

\begin{figure}[p]\ContinuedFloat
\centering
\tasksamplefig{3}
\caption[]{\textbf{(c)} telecom.}
\end{figure}

\clearpage

\section{Prompts}
\label{app:prompts}

This appendix contains the prompt templates used in each pipeline stage, reproduced verbatim from the running implementation. All prompts are domain-agnostic: the same template is used across the airline, retail, and telecom domains, with the domain environment (policy document, tool specification, schema) supplied through the templated placeholders. Placeholders of the form \texttt{\{name\}} are substituted with the corresponding runtime value described in Appendix~\ref{app:method_implementation}; the double braces \texttt{\{\{} and \texttt{\}\}} are Python format-string escapes that render as single braces in the text sent to the LLM and are otherwise passed through unchanged.

\promptlisting{RoyalBlue}{Prompt for \texttt{validate\_action\_sequence}}{assets/prompts/validate_action_sequence.txt}{box:validate-action-sequence}

\promptlisting{ForestGreen}{Prompt for \texttt{create\_user\_task}}{assets/prompts/create_user_task.txt}{box:create-user-task}

\promptlisting{ForestGreen}{Prompt for \texttt{generate\_db\_initialization}}{assets/prompts/generate_db_initialization.txt}{box:generate-db-init}

\promptlisting{ForestGreen}{Prompt for \texttt{task\_coherence\_review}}{assets/prompts/task_coherence_review.txt}{box:task-coherence-review}

\promptlisting{Plum}{Prompt for \texttt{adversarial\_strategy}}{assets/prompts/adversarial_strategy.txt}{box:adv-strategy}

\promptlisting{Plum}{Prompt for \texttt{adversarial\_scenario}}{assets/prompts/adversarial_scenario.txt}{box:adv-scenario}

\begingroup
\renewcommand{\answerYes}[1][]{[Yes]#1}
\renewcommand{\answerNo}[1][]{[No]#1}
\renewcommand{\answerNA}[1][]{[N/A]#1}
\section*{NeurIPS Paper Checklist}

\begin{enumerate}

\item {\bf Claims}
    \item[] Question: Do the main claims made in the abstract and introduction accurately reflect the paper's contributions and scope?
    \item[] Answer: \answerYes{}
    \item[] Justification: The abstract and introduction state three contributions: (i) formalizing the required desiderata for tool-using-agent benchmarks, (ii) a tool-sequence-first pipeline (TASTE) that targets each of those criteria, and (iii) an instantiation on three $\tau$-Bench domains, with an additional AppWorld adaptation. The desiderata are formalized in Section~\ref{sec:benchmark_quality}, the pipeline in Section~\ref{sec:method}, and the quantitative criteria results are reported in Section~\ref{sec:results} and Table~\ref{tab:main-results}.  Section~\ref{sub:appworld} explicitly qualifies the AppWorld results.

\item {\bf Limitations}
    \item[] Question: Does the paper discuss the limitations of the work performed by the authors?
    \item[] Answer: \answerYes{}
    \item[] Justification: We discussed the limitations of the work in the Method section (Section~\ref{sec:method} ) as well as in the Results section (Section~\ref{sec:results}). 

\item {\bf Theory assumptions and proofs}
    \item[] Question: For each theoretical result, does the paper provide the full set of assumptions and a complete (and correct) proof?
    \item[] Answer: \answerNA{}
    \item[] Justification: The paper does not include theoretical results (theorems, lemmas, or formal proofs); the Adaptive Contrastive $n$-gram model in Section~\ref{sub:n_gram} is a methodological construction validated empirically rather than a theoretical claim.

    \item {\bf Experimental result reproducibility}
    \item[] Question: Does the paper fully disclose all the information needed to reproduce the main experimental results of the paper to the extent that it affects the main claims and/or conclusions of the paper (regardless of whether the code and data are provided or not)?
    \item[] Answer:   \answerNo{}
     \item[] Justification: Section~\ref{sec:method} fully specifies the pipeline (sampling, clustering, generation, evolution); Appendix~\ref{app:method_implementation} provides the implementation details and full hyperparameter settings; and Appendix~\ref{app:prompts} contains the exact LLM prompts used at every pipeline stage. Section~\ref{sec:experiments_setup} lists the evaluated agent/user model pairs, the $\tau$-Bench harness configuration, and per-domain task counts.  Reproduction details for the supplemental experiments are incomplete.


\item {\bf Open access to data and code}
    \item[] Question: Does the paper provide open access to the data and code, with sufficient instructions to faithfully reproduce the main experimental results, as described in supplemental material?
    \item[] Answer:   \answerNo{}
     \item[] Justification: The pipeline implementation and the generated   \ourbenchmark  task sets for all three domains are provided   in the repository linked in the abstract, together with instructions to reproduce the main experimental results. Section~\ref{sec:method} describes the pipeline, Appendix~\ref{app:method_implementation} reports the full hyperparameter settings, and Appendix~\ref{app:prompts} contains the exact LLM prompts used at every stage. A link to the  github repository is also attached (see abstract). A complete code and data release for the supplemental experiments is not yet available.

\item {\bf Experimental setting/details}
    \item[] Question: Does the paper specify all the training and test details (e.g., data splits, hyperparameters, how they were chosen, type of optimizer) necessary to understand the results?
    \item[] Answer: \answerYes{}
    \item[] Justification: Section~\ref{sec:method} states the key hyperparameters; Appendix~\ref{app:method_implementation} reports the full hyperparameter settings for every pipeline stage and Appendix~\ref{app:prompts} contains the LLM prompts. Section~\ref{sec:experiments_setup} specifies the evaluated agent/user model pairs, the $\tau$-Bench harness configuration (defaults), the per-domain task counts (50/114/114) and costs, and the reward signal used for each domain.  The supplemental settings are summarized in Appendices~\ref{app:direct-llm} and~\ref{app:appworld}.


\item {\bf Experiment statistical significance}
    \item[] Question: Does the paper report error bars suitably and correctly defined or other appropriate information about the statistical significance of the experiments?
    \item[] Answer: \answerNo{}
    \item[] Justification: Table~\ref{tab:main-results} reports point estimates of \passone and (for airline) \passthree without explicit error bars or confidence intervals. The \passk metric, which requires success on all $k$ independent trials, captures across-trial variability for a given agent/user pair. We did not run additional seeds per task pair due to API-cost constraints (Section~\ref{sec:experiments_setup}).  The supplemental experiments also report single-run results without error bars.


\item {\bf Experiments compute resources}
    \item[] Question: For each experiment, does the paper provide sufficient information on the computer resources (type of compute workers, memory, time of execution) needed to reproduce the experiments?
    \item[] Answer:   \answerNo{}
     \item[] Justification:   The $\tau$-Bench  experiments are run through external LLM APIs (no local GPU training), so the relevant resource is API cost. Section~\ref{sec:experiments_setup} reports per-stage cost estimates for benchmark creation (per domain, per generator model) and per-task evaluation costs for each evaluated model, together with totals across the 50/114/114 task sets, and notes additional ablation runs not reported in the main results. Costs for the supplemental experiments are not reported.

\item {\bf Code of ethics}
    \item[] Question: Does the research conducted in the paper conform, in every respect, with the NeurIPS Code of Ethics \url{https://neurips.cc/public/EthicsGuidelines}?
    \item[] Answer: \answerYes{}
    \item[] Justification: We have reviewed the NeurIPS Code of Ethics and the research conforms with it. The paper involves no human subjects, no personally identifiable or scraped data, and no prohibited application areas; all data is synthetically generated within the public $\tau$-Bench customer-service domains   and AppWorld.

\item {\bf Broader impacts}
    \item[] Question: Does the paper discuss both potential positive societal impacts and negative societal impacts of the work performed?
    \item[] Answer: \answerNA{}
    \item[] Justification: This paper contributes evaluation methodology and a derived benchmark, not a deployed model or capability-increasing system. Better benchmarks generally promote more reliable and trustworthy tool-using agents (positive impact); any path to negative downstream applications is indirect and generic, no different from typical evaluation work, and we therefore do not include a dedicated broader-impacts discussion.

\item {\bf Safeguards}
    \item[] Question: Does the paper describe safeguards that have been put in place for responsible release of data or models that have a high risk for misuse (e.g., pre-trained language models, image generators, or scraped datasets)?
    \item[] Answer: \answerNA{}
    \item[] Justification: The paper releases only synthetic benchmark tasks built within the existing public $\tau$-Bench customer-service domains  and AppWorld; no pre-trained models, scraped data, or other high-misuse-risk assets are released, so no safeguards are required.

\item {\bf Licenses for existing assets}
    \item[] Question: Are the creators or original owners of assets (e.g., code, data, models), used in the paper, properly credited and are the license and terms of use explicitly mentioned and properly respected?
    \item[] Answer: \answerYes{}
    \item[] Justification: All existing assets are cited at first use: $\tau$-Bench~\citep{taubench_24_yao}, $\tau^2$-Bench~\citep{tau2barres}, and $\tau$-Bench Verified~\citep{saber2025cuadron} are released by their authors under the MIT License and we use them within its terms; the evaluated LLMs (Gemini-2.5-flash, Gemini-3-flash, Gemini-3-pro, GPT-5.2, Claude-Sonnet-4.6, DeepSeek-3.1,  DeepSeek-V4-Flash,  Qwen-32B) are accessed through their providers' public APIs under each provider's terms of service. AppWorld~\mbox{
\citep{appworld24trivedi} }\hskip0pt
uses Apache 2.0, with encrypted public redistribution required for its protected data and derivatives.

\item {\bf New assets}
    \item[] Question: Are new assets introduced in the paper well documented and is the documentation provided alongside the assets?
    \item[] Answer: \answerYes{}
    \item[] Justification: The generated   \ourbenchmark  task sets and the pipeline implementation are released   in the repository linked in the abstract, together with documentation.

\item {\bf Crowdsourcing and research with human subjects}
    \item[] Question: For crowdsourcing experiments and research with human subjects, does the paper include the full text of instructions given to participants and screenshots, if applicable, as well as details about compensation (if any)?
    \item[] Answer: \answerNA{}
    \item[] Justification: The paper does not involve crowdsourcing or research with human subjects; all task generation and validation is performed by LLMs and rule-based checks, and the only manual step is internal inspection by the authors of the 15 tasks on which no evaluated agent-user pair succeeded.

\item {\bf Institutional review board (IRB) approvals or equivalent for research with human subjects}
    \item[] Question: Does the paper describe potential risks incurred by study participants, whether such risks were disclosed to the subjects, and whether Institutional Review Board (IRB) approvals (or an equivalent approval/review based on the requirements of your country or institution) were obtained?
    \item[] Answer: \answerNA{}
    \item[] Justification: The paper does not involve crowdsourcing or research with human subjects, so IRB approval is not applicable.

\item {\bf Declaration of LLM usage}
    \item[] Question: Does the paper describe the usage of LLMs if it is an important, original, or non-standard component of the core methods in this research? Note that if the LLM is used only for writing, editing, or formatting purposes and does \emph{not} impact the core methodology, scientific rigor, or originality of the research, declaration is not required.
    \item[] Answer: \answerYes{}
    \item[] Justification: LLMs are core components of the proposed method (TASTE): they serve as plausibility judges during $n$-gram training and clustering validation, as generators of scenarios, databases, and user instructions, as the hint-assisted verifier agent, and as the strategy/perturbation/scenario-rewriting modules in adversarial evolution. The specific models used at each stage are listed in Section~\ref{sec:experiments_setup}, and the exact prompts are provided in Appendix~\ref{app:prompts}.  The direct baseline uses GPT-6 Luna through Codex (Appendix~\ref{app:direct-llm}), and AppWorld model roles are described in Appendix~\ref{app:appworld}.


\end{enumerate}

\endgroup

\end{document}